# A Circle Grid-based Approach for Obstacle Avoidance Motion Planning of Unmanned Surface Vehicles

Man Zhu[a,b], Changshi Xiao[a,b,c], Shangding Gu[c*], Zhe Du[c†], Yuanqiao Wen[a,b]

a. Intelligent Transportation System Research Center, Wuhan University of Technology, Wuhan, 430063, China;
b. National Engineering Research Center for Water Transport Safety, Wuhan, 430063, China;
c. School of Navigation, Wuhan University of Technology, Wuhan, 430063, China.

**Abstract:** Aiming at an obstacle avoidance problem with dynamic constraints for Unmanned Surface Vehicle (USV), a method based on Circle Grid Trajectory Cell (CGTC) is proposed. Firstly, the ship model and standardization rules are constructed to develop and constrain the trajectory, respectively. Secondly, by analyzing the properties of the circle grid, the circle grid tree is produced to guide the motion of the USV. Then, the kinematics and dynamics of the USV are considered through the on-line trajectory generator by designing a relational function that links the rudder angle, heading angle, and the central angle of the circle grid. Finally, obstacle avoidance is achieved by leveraging the on-line trajectory generator to choose a safe, smooth, and efficient path for the USV. The experimental results indicate that the proposed method can avoid both static and dynamic obstacles, have better performance in terms of distance cost and steering cost comparing with the related methods, and our method only takes 50% steering cost of the grid-based method; the collision avoidance path not only conforms to the USV dynamic characteristic but also provides a reference of steering command.

---

*Corresponding Author. E-mail address: gshangd@163.com

*Corresponding Author. E-mail address: duzhe9224@foxmail.com

# 1. Introduction

For an unmanned surface vehicle (USV), obstacle avoidance is always the key to make sure the various maritime missions are carried out (Li et al., 2021). Since the unmanned vehicles are often underactuated systems, such as Unmanned Ground Vehicle (UGV), Unmanned Aerial Vehicle (UAV) and Unmanned Surface Vehicle (USV), obstacle avoidance operation should take into consideration the kinematics and dynamics constraints of the vehicles (Bullo et al., 2001; Belta et al., 2005; Wei-Dong et al., 2020; Zhou et al., 2020a). Thus, the USV navigation problem for obstacle avoidance considering kinematics and dynamics constraints is discussed in this paper.

In general, there are two problems in obstacle avoidance that need to be settled: map (environment) modeling and avoidance algorithm.Map modeling is the foundation of the avoidance algorithm. Without a proper and accurate map representation, the avoidance algorithm cannot give a full play. There are three ways to express the environment: Metric Representation, Topological Representation (Zhou et al., 2020b), and Hybrid Representation (Filliat et al., 2003). Metric Representation uses a world coordination system to express environmental characteristics. It is usually classified into Spatial Decomposition and Geometric Representation. Grid Map (Lee et al., 2011) and Quad-Tree (Yahja et al., 1998) belong to the former, Visibility Graph (Niu et al., 2020) belongs to the latter. Topological Representation uses nodes to express the location and sides to express the relation of these nodes. The typical method is the Voronoi Graph (Wang et al., 2020). Hybrid Representation is the way of combining metric information and topological characteristic. The typical way is to extract topological characteristics from the metric map, which is called Simultaneous Location and Mapping, SLAM (Thrun et al., 1998).

For the avoidance algorithm, there are plenty of methods proposed for vehicle collision avoidance. For example, Artificial Potential Field (APF) (Orozco-Rosas et al., 2019; Lyu et al., 2019) and Velocity Obstacle (VO) (Tan et al., 2020; Chen et al., 2020a; Chen et al., 2020b) are the most widely used methods. The former takes advantage of the virtual potential to make research objects keep away from the obstacles, the latter makes use of relative velocity to avoid moving obstacles. Genetic Algorithm (Mirjalili, 2019), Neural Network Algorithm (Wu et al., 2019; Woo et al., 2020), etc. are the intelligent algorithms that use the idea of biology. When the dynamic constraints of the research object are considered, some special curves like Dubins Path (Jha et al., 2020), Fermat Spiral (FS) (Dahl, 2013), B-Spline curve (Choi et al., 2017) and Clothoid curve (Shanmugavel et al., 2010) are often combined with the above algorithms to make the path of collision avoidance more smooth and continuous.

Different from the above methods, a motion planning method for the USV obstacle avoidance based on Circle Grid Trajectory Cell (CGTC) considers the kinematics and dynamics constraints and can provide a smooth manipulation and safety path during the USV obstacle avoidance. The designed Circle Grid allows the USV's possible headings and waypoints to have a continuous change, which is related to the continuous motion curve. Meanwhile, the structure of the circle grid tree can fully cover the planning space. The Trajectory Cell is built by the proposed standardization rules. These rules discretize trajectories and maintain the continuity. Combining the Circle Grid and the Trajectory Cell, the CGTC is introduced to realize the smooth manipulation during obstacle avoidance.

Overall, the contributions of our proposed method are as follows:

1. The map of the searching space is built based on the circle grid to make sure the possible headings and waypoints of the USV have a continuous change in the searching process.

2. The searching method is designed considering kinematics and dynamics constraints by blending in the mathematical model of the USV, which makes the planned path conform to the actual motion of the USV

3. The obstacle avoidance strategy is proposed based on the CGTC scheme that works in both static and dynamic environments, and the manipulation during the USV collision avoidance process is guaranteed to be smooth.

## 2. Related work

The traditional avoidance methods for an USV often combine classical algorithms with marine instruments or regulations. For example, Zhuang et al. (2006) combine electronic charts (e-chart) and marine radar with the Dijkstra algorithm to achieve global and local route planning for USV. They use e-chart to deal with global planning and use a smoothing processed radar image (binary map) to deal with local collision avoidance. Almeida et al. (2009) take advantage of marine sensors to detect obstacles. They use a camera to get the environment information and grade the degree of danger based on the distance of the obstacle. International Marine Collision Regulations (COLREGS) also play an important role in the maritime field, which all the ships in the sea must obey (Han et al., 2020). So the regulations are introduced to normalize the process of collision avoidance. Some scholars (Campbell, 2012; Yang, 2016) firstly plan a global path by A* algorithm. Then based on COLREGS, they classify the collision into different situations and work out different behaviors to avoid each situation. Those behaviors correspond to different routes. Sing, Y. (2019) and Mina, T., et al. (2019) propsed a safe A* algorithm for USV safety, where the distance between USVs and obstacle is considered in sea surface currents. Song et al. (2019) introduce a smooth A* algorithm for USV path planning and collision avoidance, where the method can provide a more continuous route and have no redundant path points by reducing the unnecessary jags.

When dynamic constraints are considered, there are different obstacle avoidance methods for different kinematic conditions. Curve Fitting is a way of dealing with the constraint of the minimum turning radius and the heading. The main idea is to construct a curve to make the best fit for a certain polyline path and meet the headings of the origination and destination. Chen (2016) combines the "CLC" Dubins path with a genetic algorithm to improve the optimization of the pose (positions and heading angles) transition from the starting point to the end point. Sun (2016) takes the floating-point numbers and the turning radius of KT equations as the order and the radius of curvature of the Bessel curve to restrain the collision avoidance path, and gets a smooth continuous trajectory for USV. When dealing with the constraint of the speed, Velocity Obstacle is usually used. Kuwata et al. (2011) firstly build a coordinate system of velocity space. Then the overlapping area that the USV may have a collision in this system is calculated. Finally, by combining with the COLREGS, the collision avoidance direction of the USV is judged. Du et al. (2015) combine Velocity Obstacle with Particle Swarm Algorithm to calculate the optimal heading and speed for obstacle avoidance. Compared to the above-mentioned methods, Multi-constraint optimization is more direct. It treats these dynamic constraints as objects that need to be optimized.

Moreover, Kim et al. (2014) take into account the possible heading angles to turn the original two-dimensional ($x$, $y$) planning space into three-dimensional ($x$, $y$, $\theta$). Yang et al. (2015) take advantage of the angle constraint to improve the A* algorithm and consider the ship dimension and obstacle security boundary in the collision avoidance process. Huang et al. (2020a) introduce a framework of human-machine interaction for safe collision avoidance of USVs, which can provide an interpretable decision-making process for the operator, and can be suitable to the under-actuated features of USVs. Huang et al. (2020b) provide a method of collision risk assessment for USV collision avoidance, in which the USV dynamics constraints are considered during the risk assessment of time-varying USV collision avoidance.

Apart from the above methods, some scholars start to consider dynamic constraints in avoidance operations in recent. For example, Svec et al. (2012) take advantage of the idea of probability prediction. They firstly use USV mathematical model to predict all the possible collision avoidance trajectories. Then use min-max game-tree to select the smallest collision probability trajectory from the possible trajectories pool. Du et al. (2018) make use of USV's hydrodynamic model to build a set of trajectory segments. Through the location and direction that those segments can reach, they propose a search and collision avoidance strategy containing the movement characteristics of the USV. Gu et al. (2019) propose a path search method for USV path planning, in which they achieve efficient and time-saving path search that fascinates USV obstacle avoidance. And Gu et al. (2020) develop a motion planning method that improves the A* algorithm and considers USV dynamic constraints to avoid obstacles in restricted waters. Zhou et al. (2020b) provide a better performance method for USV motion planning comparing with the related algorithm, in which the method can achieve safe motion planning and avoid obstacles.

However, the above research works pay more attention to the quality of the planned path but less about the maneuverability for obstacle avoidance operation under dynamic environments. A practicle motion obstacle avoidance method for an USV not only requires improving the quality of the path but also needs to give a set of reference commands. Thus, this paper not only focuses on which path is safe but also pays attention to how to avoid these obstacles by manipulating smoothly.

# 3. CGTC Model

## 3.1 Problem Statement

Generally, the grid map method is one of the most widely used map modeling methods for USV collision avoidance. However, the grid map is unfit for associating with a continuous motion curve, because there are finite candidate nodes (discretized nodes) in every loop of search (Figure 1(a)) and can cover the area of the search probability.

To make the discretized nodes become a set of continuous nodes (curve segment), the shape of a grid should be changed. When the basic grid is circle-shaped and the candidate nodes are on the arc, the number of nodes is infinite and they form a continuous curve (Figure 1(b)). Thus, we design a circle grid map modeling method to solve the combination between the dynamic constraints and the search algorithm in the USV motion planning problem.

Moreover, Du et al. (2018) and Zhou et al. (2020b) constructed the grid map and topological map for USV motion planning using trajectory unit respectively, which can consider the combination between the dynamic constraints and the search algorithm in USV motion planning problem. Although the method of Du et al. (2018) can achieve motion planning considering dynamic constraints very well, the method may not get the short path; After that, Zhou et al. (2020b) develop a motion planning method based on the topological map and provide better performance comparing with Du et al. (2018), such as the shorter path distance and the shorter search time, nonetheless they may need to further explain and proof the effectiveness of the method using the trajectory unit, and design the sophisticated strategies for USV collision avoidance.

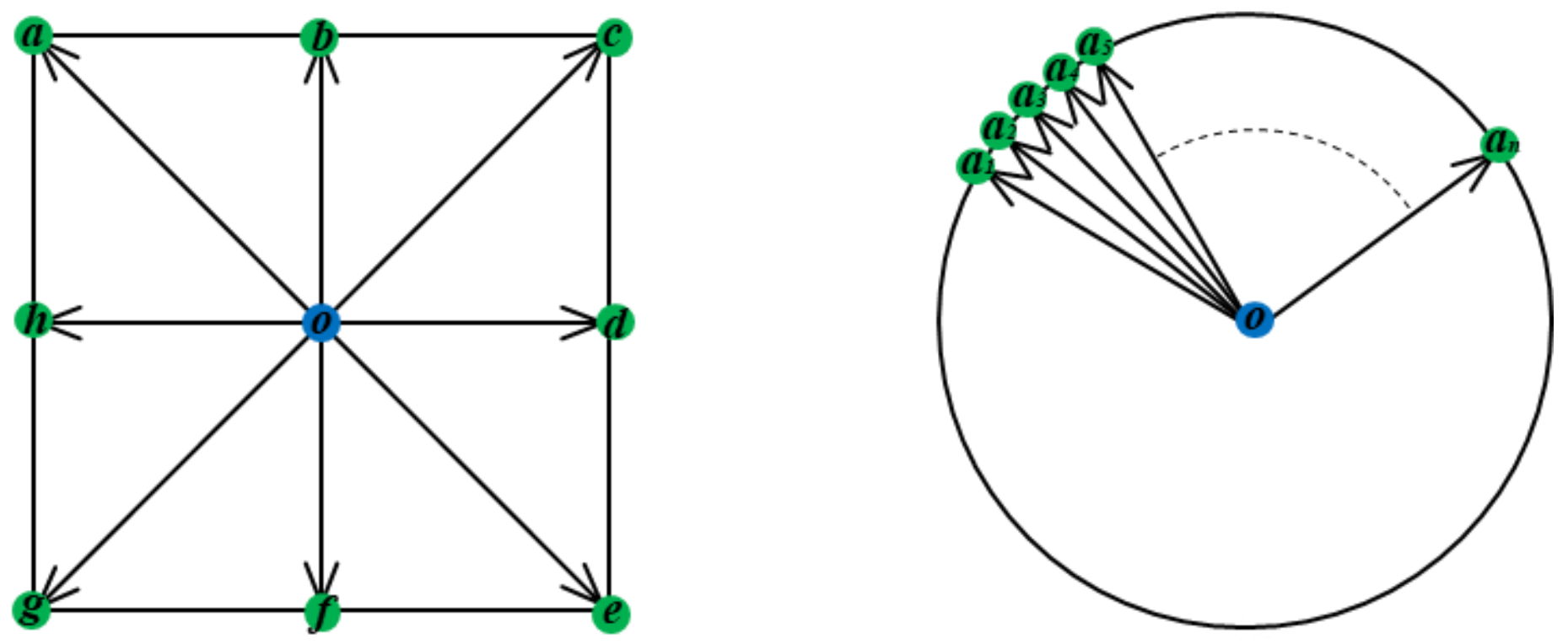

(a) (b)

**Figure 1.** The possible orientations: (a) In a grid map; (b) In a circle map.

Therefore, to solve the above problems and achieve better performance for USV collision avoidance, several steps need to be done well:

Firstly, the representation of dynamic constraints. The motion curves not only contain all the dynamic characteristics but the relationships of these constraints. Thus, the trajectory is a good way of representation, and how the trajectory is produced is the first problem.

Then, the standardization of the trajectories. Although the trajectory is a good way of representing the dynamic constraints, the final purpose is to work for collision avoidance. Thus, it is necessary to build a set of rules to standardize these trajectories. The rule must keep the trajectories simple which makes the path easy to realize, and also keep them coupled to make the final path still maintain continuously.

Finally, the coverage of the circle-shaped map. The reason why the grid map is widely used in the path planning problem is that it can cover the research space completely (Figure 2(a)). But for a circle map, there are some uncovering areas because of the arc (Figure 2(b), the white shadow). Thus, how to solve this defect is important for the map modeling and the premise of the accuracy for path searching.

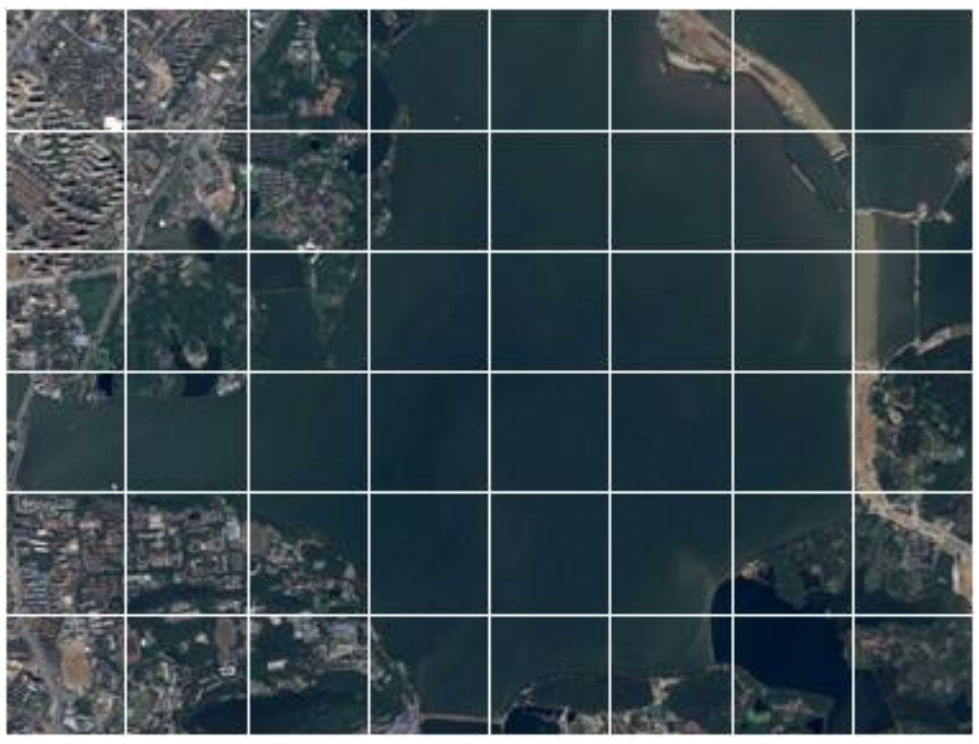

(a)

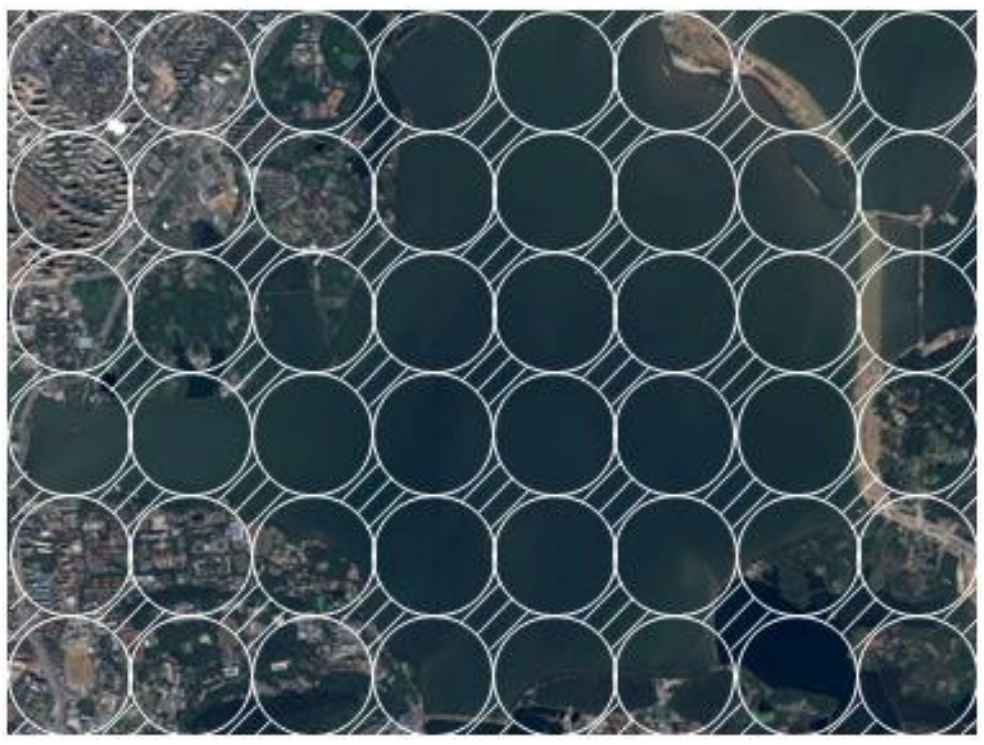

(b)

**Figure 2.** The coverage rate: (a) In grid map, the coverage is complete; (b) In circle map, there are some uncovering areas.

To achieve the above steps, this Section establishes the CGTC model. Section 3.2 introduces a ship's mathematical method to produce the USV trajectories. Section 3.3 builds a trajectory standardization rule to constraint them. Section 3.4 proposes a circle cell tree to solve the full space coverage problem. In the end, the CGTC model is established in Section 3.5.

## 3.2 Ship Mathematical Model

The ship's mathematical model is built by the Maneuvering Mathematical Group (MMG) method (Ogawa et al., 1977). The main idea is to decompose fluid forces and moment into several parts which affect the hull, propeller, and rudder; the external effect on the hull is classified into inertia and viscidity (Kijima et al., 1990). Therefore, an MMG model consists of an inertia model, a viscidity model, a propeller model, and a rudder model.

To simplify the problem, three assumptions are made as follows.

**Assumption 1.** Only planar motions are considered (neglect heaving, rolling, and pitching).

**Assumption 2.** No account of the influence of wind, current, and waves.

**Assumption 3.** The speed is considered as a constant in a fixed propeller gear when the ship is steady steaming.

Aiming at **Assumption 3**, it is necessary to make an explanation. When a ship sails on the waters, because of the resistance, the load of the propeller will increase. And it will result in the decrease of the propeller speed (*n*). At the moment, the speed controller will raise the engine power to offset the increased load of the propeller and maintain the speed. In this whole process, the engine works in the closed-loop system of the automatic rotation speed control.

Thus, the model is based on three degrees of freedom: yawing, swaying, and surging:

$$\begin{bmatrix} X \\ Y \\ N \end{bmatrix} = \begin{bmatrix} X_I + X_H + X_P + X_R \\ Y_I + Y_H + Y_P + Y_R \\ N_I + N_H + N_P + N_R \end{bmatrix} \tag{1}$$

where *I*, *H*, *P*, *R* denote the forces (or moment) of inertia, viscidity, propeller, and rudder respectively, and the detailed model can be seen in the references (Ogawa et al., 1977; Du, et al., 2018).

## 3.3 Trajectory Standardization Rule

The standardization rule is the key to the Trajectory Cell model. An appropriate rule not only builds a bridge between motion curves and avoidance algorithm but also guarantees the final trajectory continuous when these discretized trajectory segments splice each other. There are three rules to standardize the trajectory segments.

**Rule 1**: The motion states are consistent and stable at the start and end moments.

This rule is to guarantee the final trajectory continuous when these discretized trajectory segments splice each other.

The motion states of a USV include position ($x$, $y$), heading ($\theta$), velocity (advance speed $u$ and transverse speed $v$), and rudder angle ($\delta$). The first three parameters ($x$, $y$, $\theta$) are the variable in path searching, so they are always changing. And this rule is to make sure that at the start and end moments: (1) the velocities are identical; (2) the rudder angles are zero.

The first requirement is to guarantee the curvature of the adjacent segments does not change. For a ship, when the rudder is fixed, different velocities will produce a different turning radius. If the velocity is fixed, the turning radius will keep the same. The second one is to keep the course stable. When the rudder angle is zero, the USV will fix it on a certain course. So the current trajectory segment will transfer to the next smoothly (Figure 3).

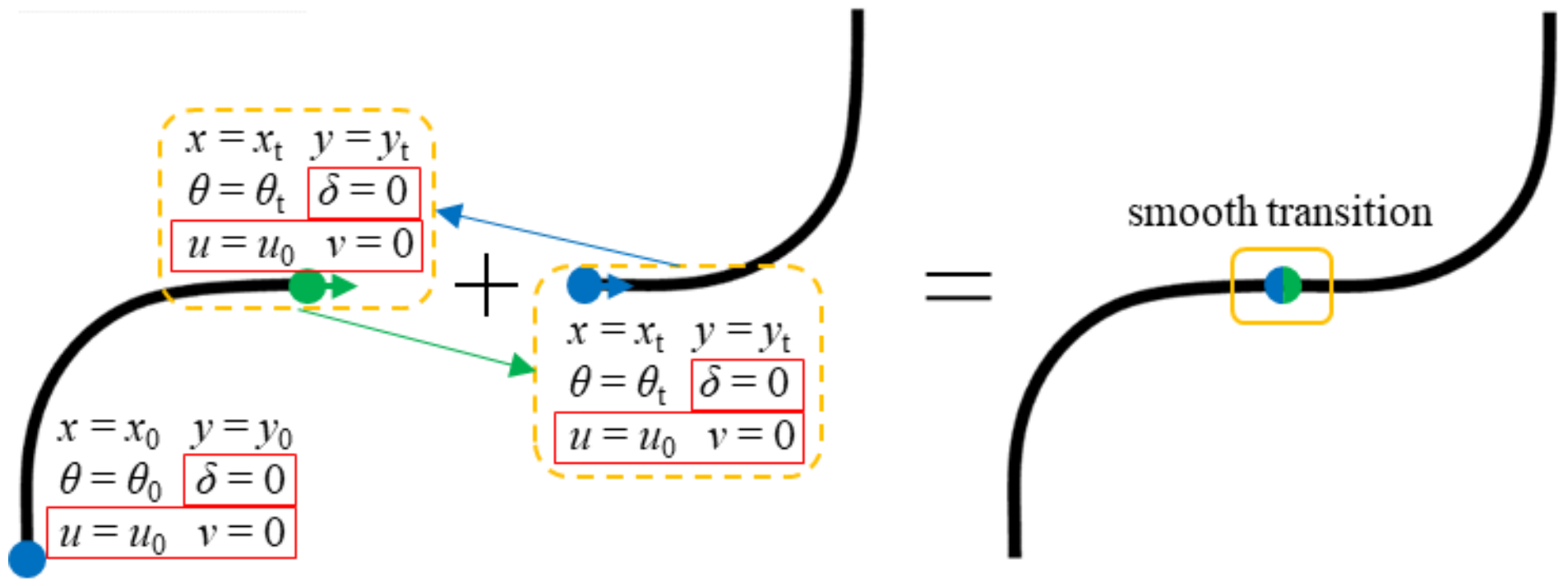

**Figure 3.** Rule 1, the motion states are consistent and stable at the start and end moments: the rudder angles are zero and the velocities are identical.

**Rule 2:** The number of steering is no more than once.

This rule is to optimize the final path and make the manipulation smooth and closer to reality.

As a basic unit composing the final trajectory, the Trajectory Cell should be as simple as possible. For the trajectory segment itself, there should not be too much steering (Figure 4 (a)); otherwise, the finial-planned trajectory will have too many inflection points, which does not accord to the real manipulation (Figure 4 (b)). Besides, according to **Rule 1**, the rudder angle of each trajectory segment will back to zero in the end. Thus, the number of steering should no more than once in a cell.

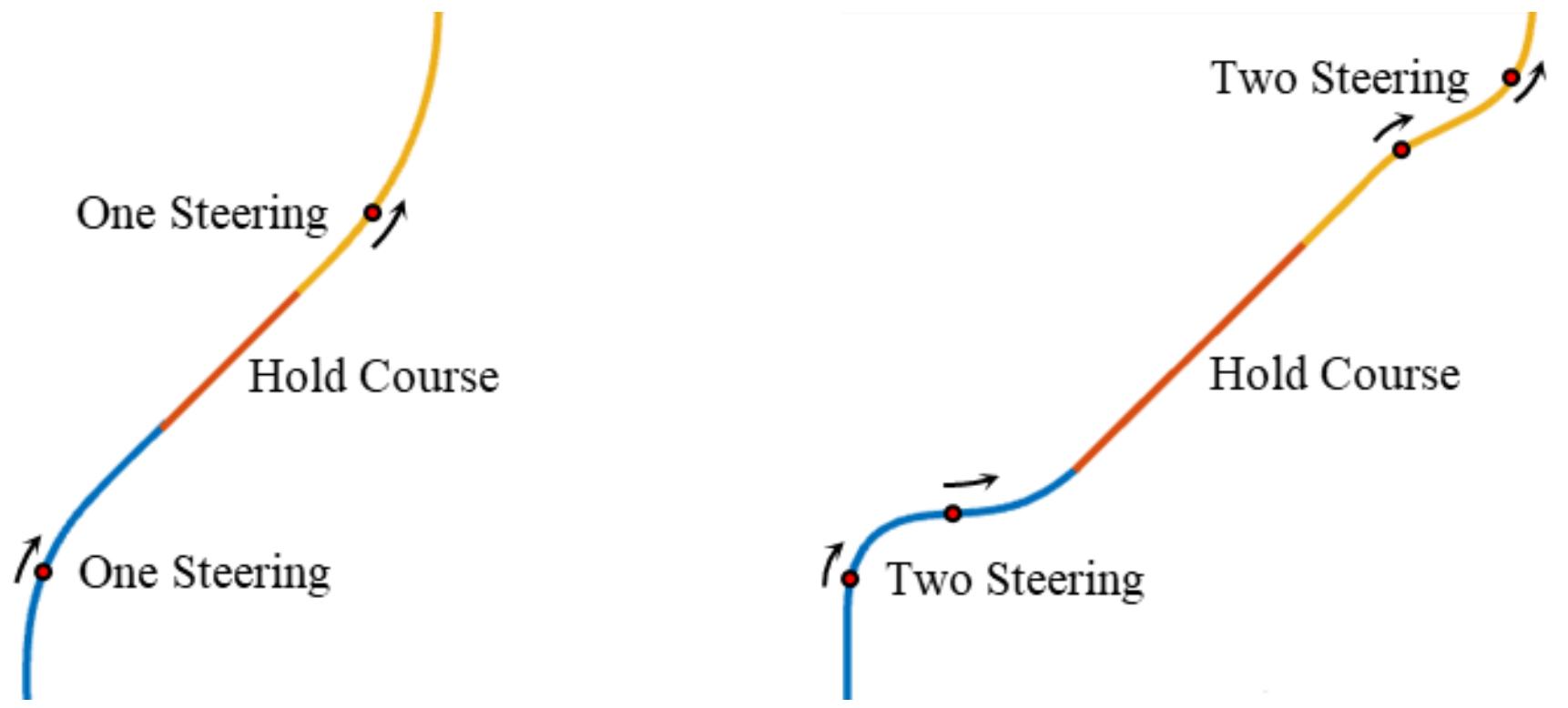

(a) (b)

**Figure 4.** Rule 2, the number of steering is no more than once: (a) one steering in a cell; (b) two steering in a cell.

**Rule 3:** The distances between the start and end point from all the cells are identical.

This rule is to make preparation for the Circle Grid model.

To make all the Trajectory Cells as a Circle Grid form to path search, the distance between the start and end point should be the same. So when the start point is fixed, all the possible end points will make up a circle arc. This arc is a candidate point set that includes all the possible USV motion states. Thus, dynamic constraints are transformed into an arc segment.

## 3.4 Circle Grid Tree

Section 3.1 mentions that the coverage of the circle-shaped map is the key for the map modeling. But there will be some uncovering areas if the circles are tangent to each other, so they must put overlapped. There are three ways of overlapping the circles: Rectangular Overlapping, Homocentric Overlapping, and Tree Overlapping.

The Rectangular Overlapping is similar to a grid map (Figure 5 (a)). The distance between two circles is the radius, and the adjacent circles cross each other's center. However, this configuration neglects many situations. It can be seen from Figure 5 (b), except for the blue points, there are no other candidate points. When a trajectory is a red curve shown in Figure 5 (b), the map cannot provide the next waypoint. Thus, this configuration is inappropriate.

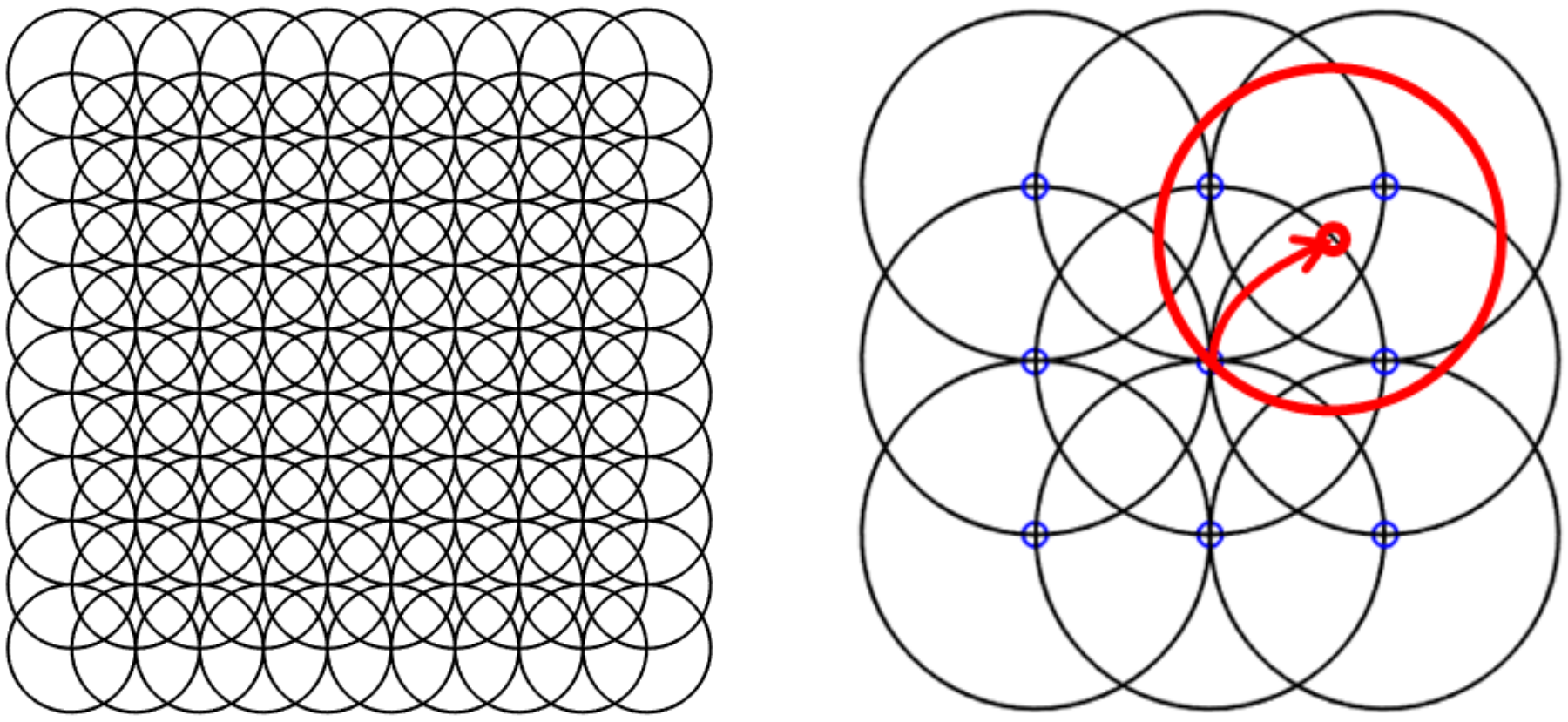

(a) (b)

**Figure 5.** Rectangular Overlapping: (a) the configuration; (b) the defect.

The Homocentric Overlapping is a set of circles with different radius and the same center (Figure 6 (a)). This configuration solves the above problem, all the situations can be found on the circle arc. And all the positions of the waypoints can be located by the polar coordinates (the center is the origin).

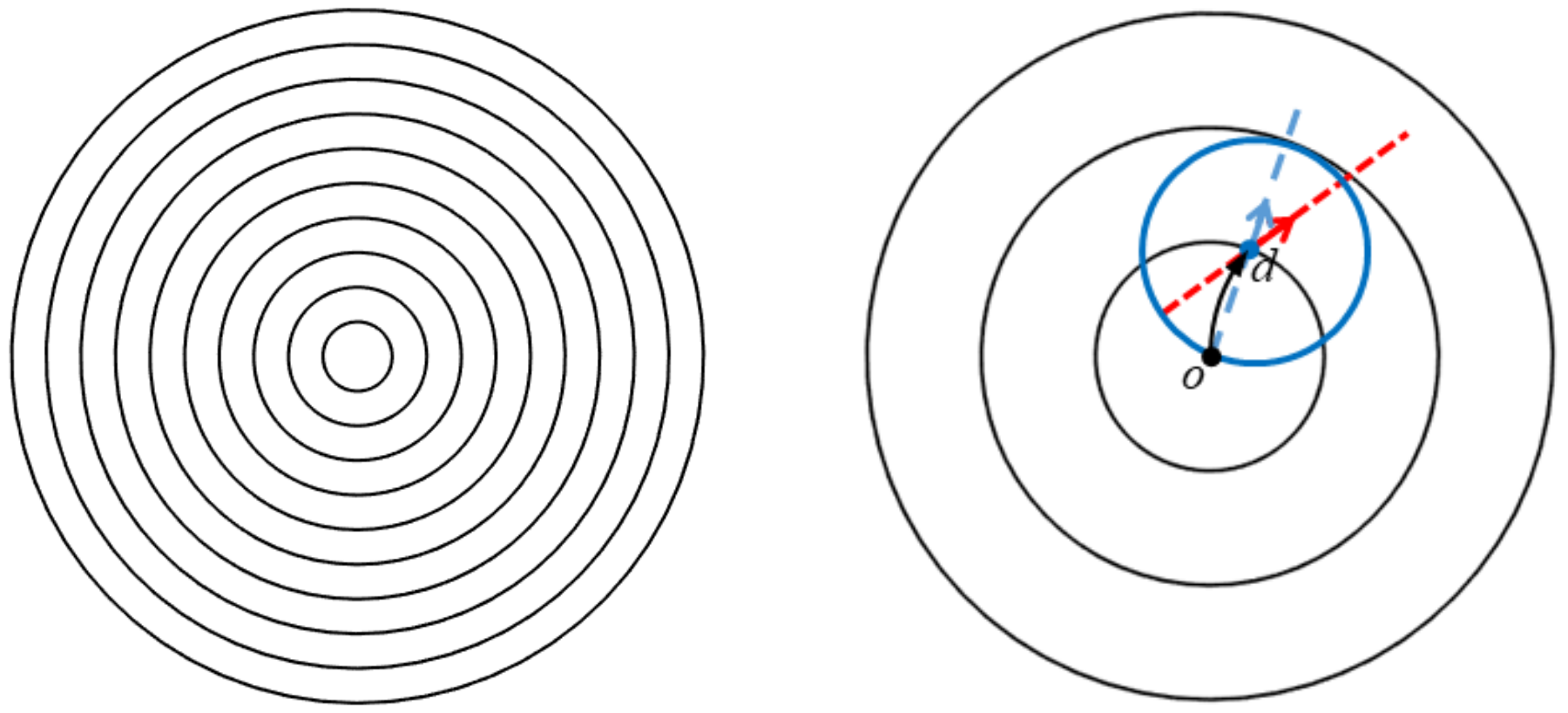

(a) (b)

**Figure 6.** Homocentric Overlapping: (a) the configuration; (b) the defect.

However, this configuration has to meet one requirement: the reverse extension line of each heading has to go through the center. It can be seen from Figure 6 (b), when a trajectory is from *o* to *d*, if the heading in position *d* is the blue arrow, the position of the next candidate points set can be calculated by using the polar coordinates. However, according to the tangent line of the trajectory, the real heading is the red arrow, and the reverse extension line of this heading is not through the center. So the next candidate points set is hard to be calculated.

Thus, this configuration has a high demand for trajectories, and it is also inappropriate.

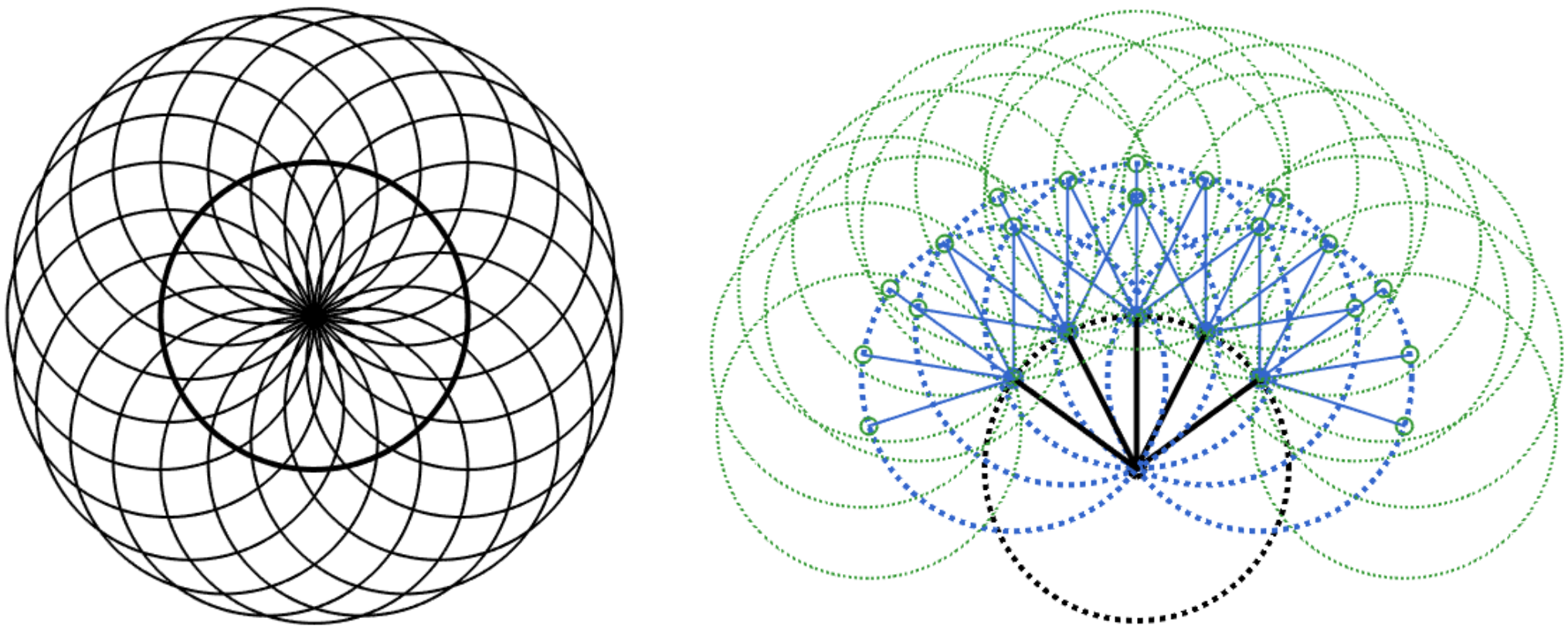

(a) (b)

**Figure 7.** Tree Overlapping: (a) the configuration; (b) the circle grid tree.

Actually, Tree Overlapping is the best choice. This configuration takes advantage of the structure of the Unit Tree. In each layer of the tree, the current point is always the center of the circle, and the next candidate points set are always located on the arc of this circle (Figure 7 (a)). So the position is calculated by ship coordinate (moving coordinate) system first, and then transform to world coordinate system.

When it is connected all the possible candidate points from the first layer to the bottom, the nodes and edges make up a tree (Figure 7 (b)). Thus, this configuration not only covers all the possible situations but also has an easy way to calculate the positions. And this paper will take advantage of it to build the CGTC model.

## 3.5 Producing the CGTC

Based on Section 3.2, 3.3, and 3.4, the CGTC model can be produced. It can be seen from Figure 8, there are three states and two stages in the process of producing CGTC.

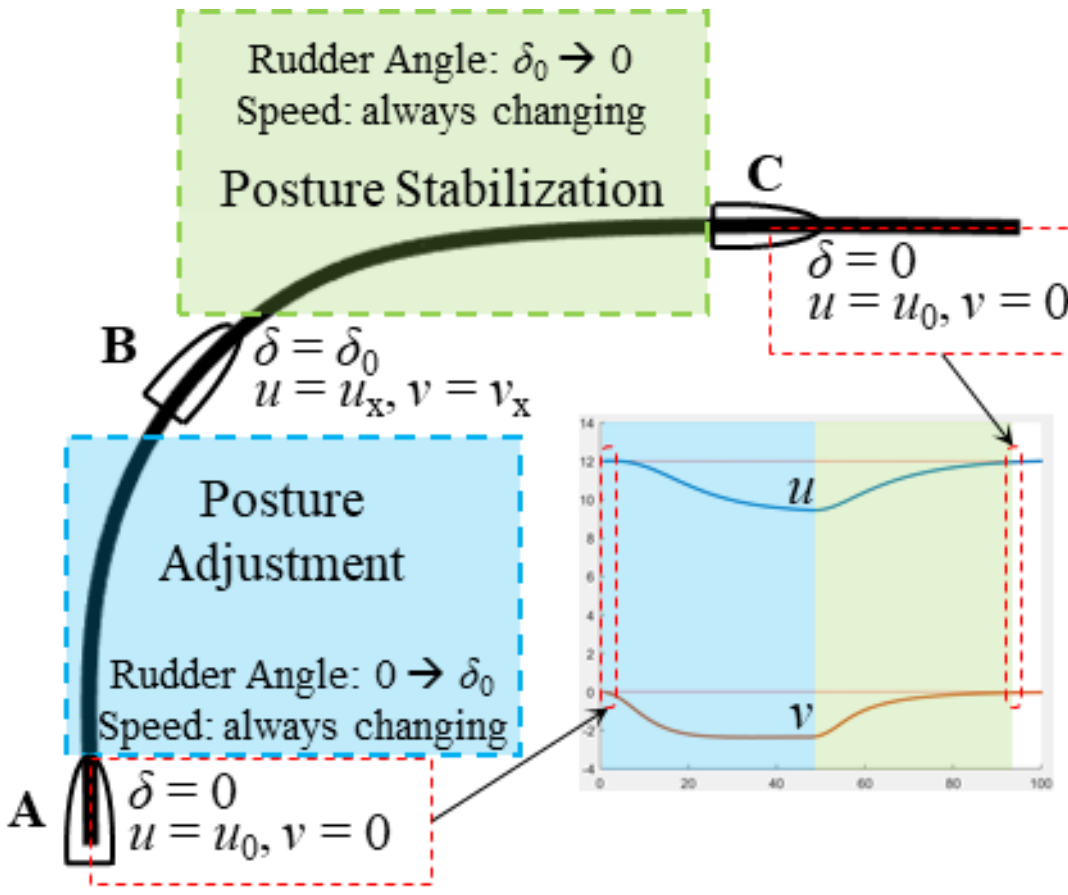

**Figure 8.** The process of producing the CGTC.

At the beginning, the rudder angle is 0 and the velocity is a stable speed under a certain propeller gear ($u$=$u_0$, $v$=0), i.e. State A. The first stage is Posture Adjustment, the rudder angle is changed from 0 to a certain value $\delta_0$. In this stage, the rudder angle is fixed on $\delta_0$, the USV is on a steering process. When it is on state B, the adjustment ends. However, the current state is unstable, since the rudder angle is still changing. So the USV goes to the second stage, Posture Stabilization. In this stage, the rudder angle is changed from a certain value $\delta_0$ to 0 and remains unchanged. The USV is gradually on a stable process. After a period of time, the USV is finally on State C that the rudder angle is 0 and the velocity is the same as State A (According to **Assumption 3** in Section 3.2). The velocity-changing curve can also be seen in Figure 7. It is obvious that the velocity is always changing in the whole process. However, in State A and C, the values are the same.

Thus, according to different goal headings and taking 90° as the maximum heading change, the CGTC set can be produced by the above process. Although the heading changes on the port and starboard side are the same, the absolute value of rudder angles are different (35° and 31°). It results from the real characteristics of the ship.

Generally, the mass of a ship cannot be strictly balanced on the port and starboard side. So the center of gravity is not on the centerline. When sailing on the sea, the effect of the rudder is different between the port and starboard side. Figure 9 is the two sides turning circles under the same rudder angle produced by the ship mathematical model from section 3.2. It is obvious that the turning radius on the port side is a little larger than starboard ($R_P$> $R_S$). Thus, the rudder effect on the starboard side of the USV studied in this paper is better than the port side.

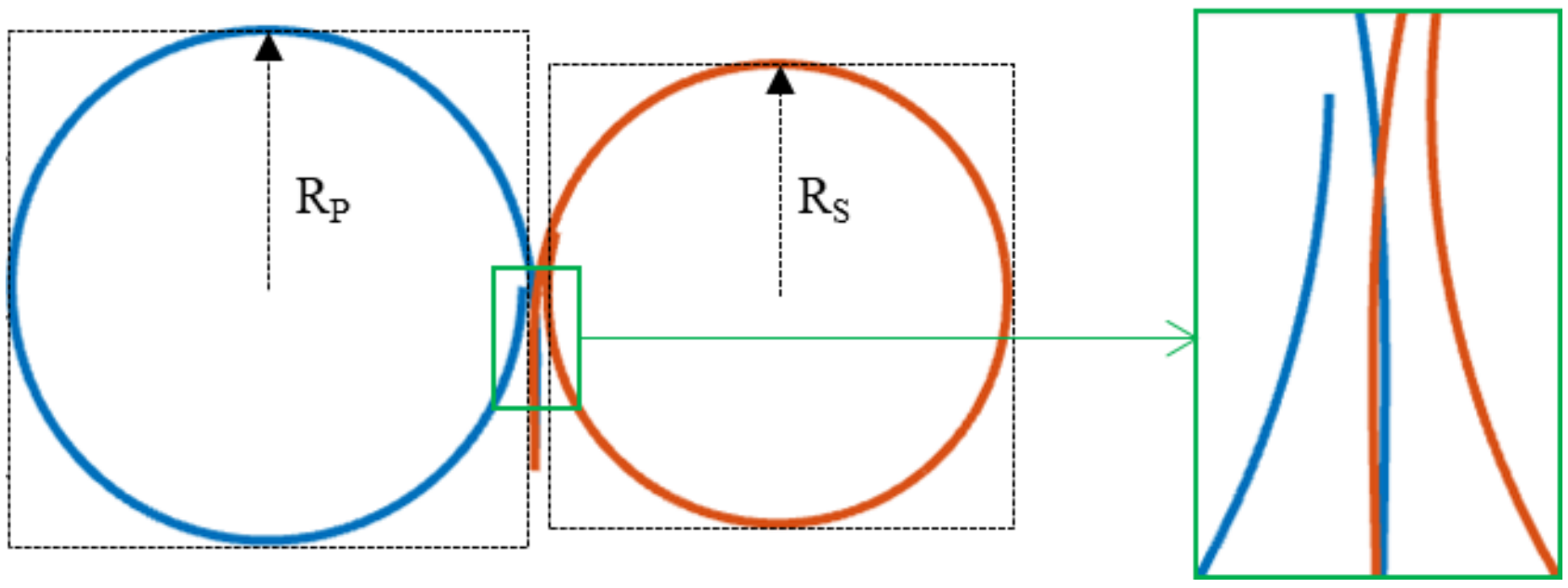

**Figure 9.** The turning circles of the USV: The steering radius on the port side is a little larger than the starboard ($R_P$> $R_S$); the green box on the right is the reverse displacement (the "kick").

In addition, it is also noticed that there is a reverse displacement (so-called "kick", the green box on the right) at the beginning of the steering. The "kick" is a transverse distance from the center of gravity to the opposite side of the steering. It is a common phenomenon that happens at the beginning of the steering. Thus, the reverse displacement confirms the accuracy of the mathematical model.

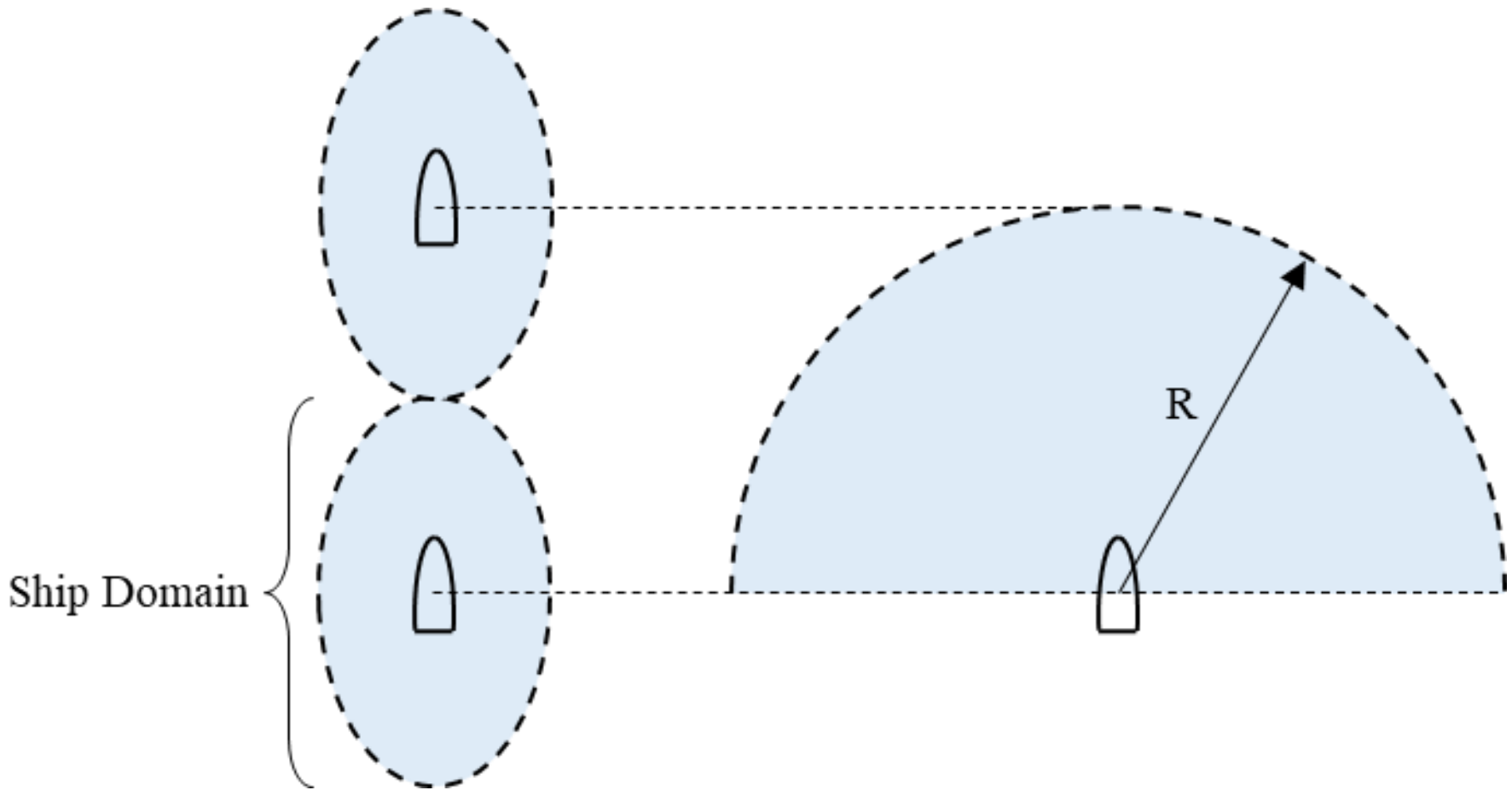

**Figure 10.** The determining of the radius of the circle grid: taking advantage of Ship Domain.

As for the radius of the circle grid, this paper adopts the concept of the Ship Domain to determine safety. The Ship Domain is a safety zone for protecting a ship from collisions from other ships. According to some researches (Fujii et al., 1971; Hansen et al., 2013), the ship domain is usually an ellipse whose major axis is 4 to 8 times the length of the ship. Thus, the radius of the circle grid is the size of the Ship Domain for the USV (shown in Figure 10).

# 4. Method of USV Obstacle Avoidance

Section 3 is Map and CGTC Modeling, and Section 4 is Planning Algorithm. This Section focuses on the strategy of path searching.

The main idea of this Section is to take advantage CGTC model to plan out the best motion curve in each search loop for the USV. Firstly, Section 4.1 designs an on-line trajectory generator for expressing the result of motion planning. The generator takes the motion parameters produced in the process of searching as the basic inputs to generate the motion curve in real-time. Then, Section 4.2 is the key to parameter solving. The three angles in the CGTC model are picked out to determine the position, heading, and manipulation of the USV. Next, aiming at free and obstacle waters, Section 4.3 and 4.4 introduce the strategy of optimal searching respectively. Finally, the flow of the algorithm is provided to connect the whole algorithm in a series.

## 4.1 On-line Trajectory Generator

The desired result of motion planning is to plan out a practical motion curve. So the trajectory generator is a necessary link in the whole algorithm, especially in the process of path searching. Thus, it is important to design an on-line trajectory generator. Before designing the generator, the factors that affect the motion curve are needed to figure out. According to the mathematical model of the USV, when the speed is fixed, three factors can affect the characteristic of the motion curve: the rudder angle ($\delta$), the position ($x$ and $y$), and the heading ($\theta$).

It can be seen from Figure 11 (a), the rudder angle determines the shape of the curve. The larger the angle, the better the rudder effect, which means a smaller turning circle. So the rudder angle is related to the radius of the curve. Figure 11 (b) is the effect of position. This factor changes the initial position of the curve and makes it have a translation transformation. Figure 11 (c) is the heading influence. Different initial headings make the USV has different reference directions which can prompt the curve to have a rotation transformation.

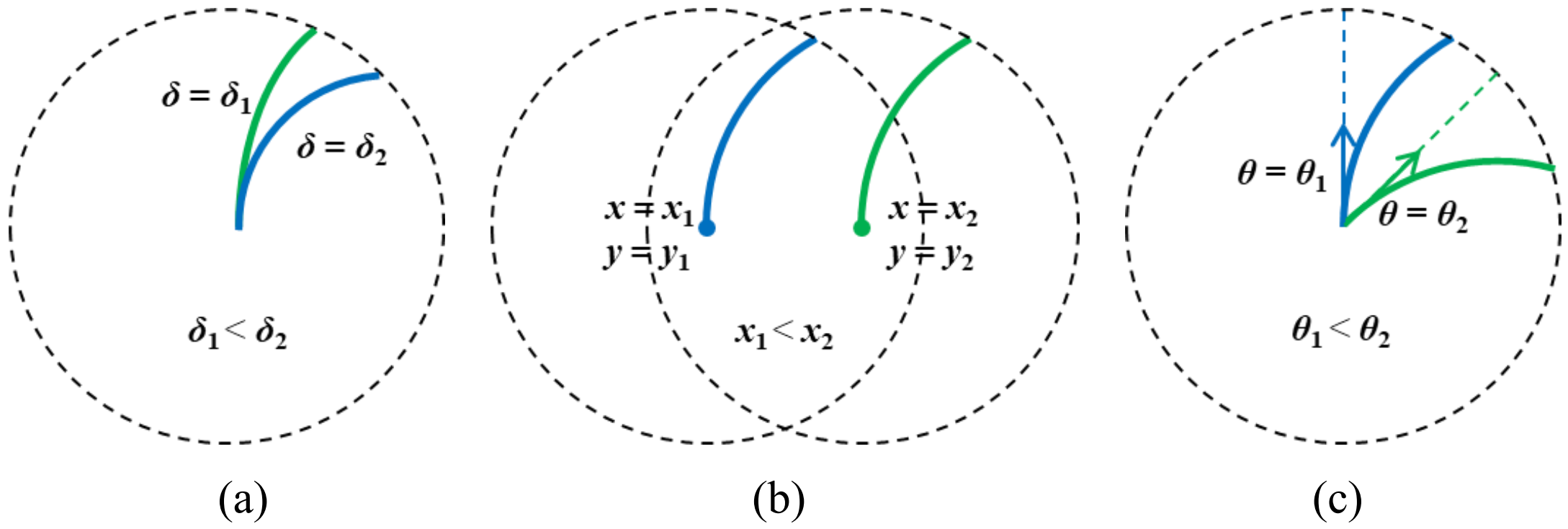

**Figure 11.** The three factors result in different motion curves: (a) the rudder angle; (b) the initial position; (c) the initial heading.

Thus, according to the above analysis, the inputs of the generator are $\delta$, $x$, $y$, and $\theta$

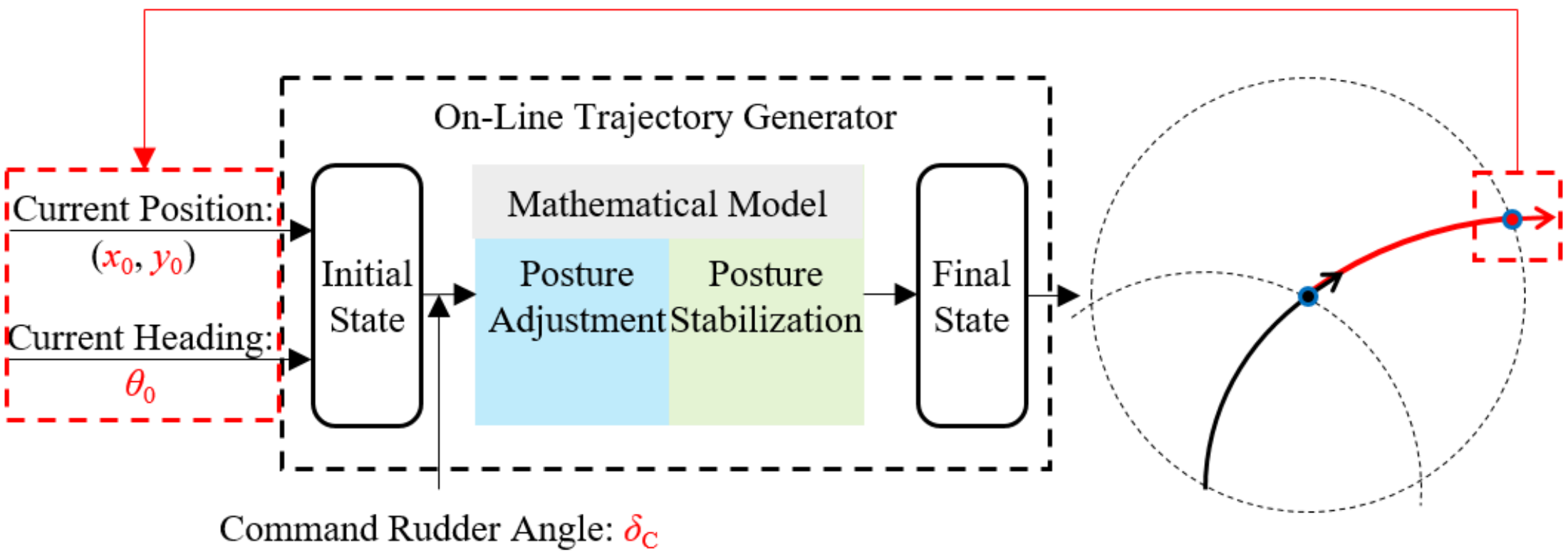

**Figure 12.** The on-line trajectory generator.

It can be seen that the position and heading are the iterative data. In each searching loop, the output data from the last loop will renew the input data of the current loop. So as long as the original motion planning state of the USV is determined, these three inputs ($x$, $y$, and $\theta$) can be easily calculated. However, as the decision input for the shape of the motion curve, the determining of the rudder angle is the most important for this generator. Thus, how to calculate the rudder angle is the next discussed problem.

## 4.2 Relational Function of Rudder, Heading and Central Angle

The way to solve the rudder angle is to find the relation between it and other parameters. According to CGTC in Section 3, there is a certain functional relationship between the rudder angle and heading.

Figure 13 (a) is the data of different rudder angles corresponding to different headings, and Figure 13 (b) is the coordinate point by these data. It can be seen that the heading has a positive relationship with the rudder angle. To verify the correctness of this relation, the correlation coefficient is calculated. The correlation coefficient is the measurement of the linear correlation between variables. The formula can be expressed as:

$$r(X,Y) = \frac{Cov(X,Y)}{\sqrt{Var[X] \cdot Var[Y]}} \tag{2}$$

where $Cov(X, Y)$ is the covariance of $X$ and $Y$, $Var[X]$ and $Var[Y]$ are the variances of $X$ and $Y$ respectively.

| Rudder ($\delta$) | Heading ($\theta$) |
|---|---|
| -35 | -89.2433 |
| -29 | -80.761 |
| -23 | -69.3535 |
| -17 | -54.8467 |
| -11 | -37.0651 |
| -5 | -17.0874 |
| 1 | 3.8501 |
| 7 | 26.1262 |
| 13 | 46.9387 |
| 19 | 64.3923 |
| 25 | 78.3661 |
| 31 | 89.3798 |

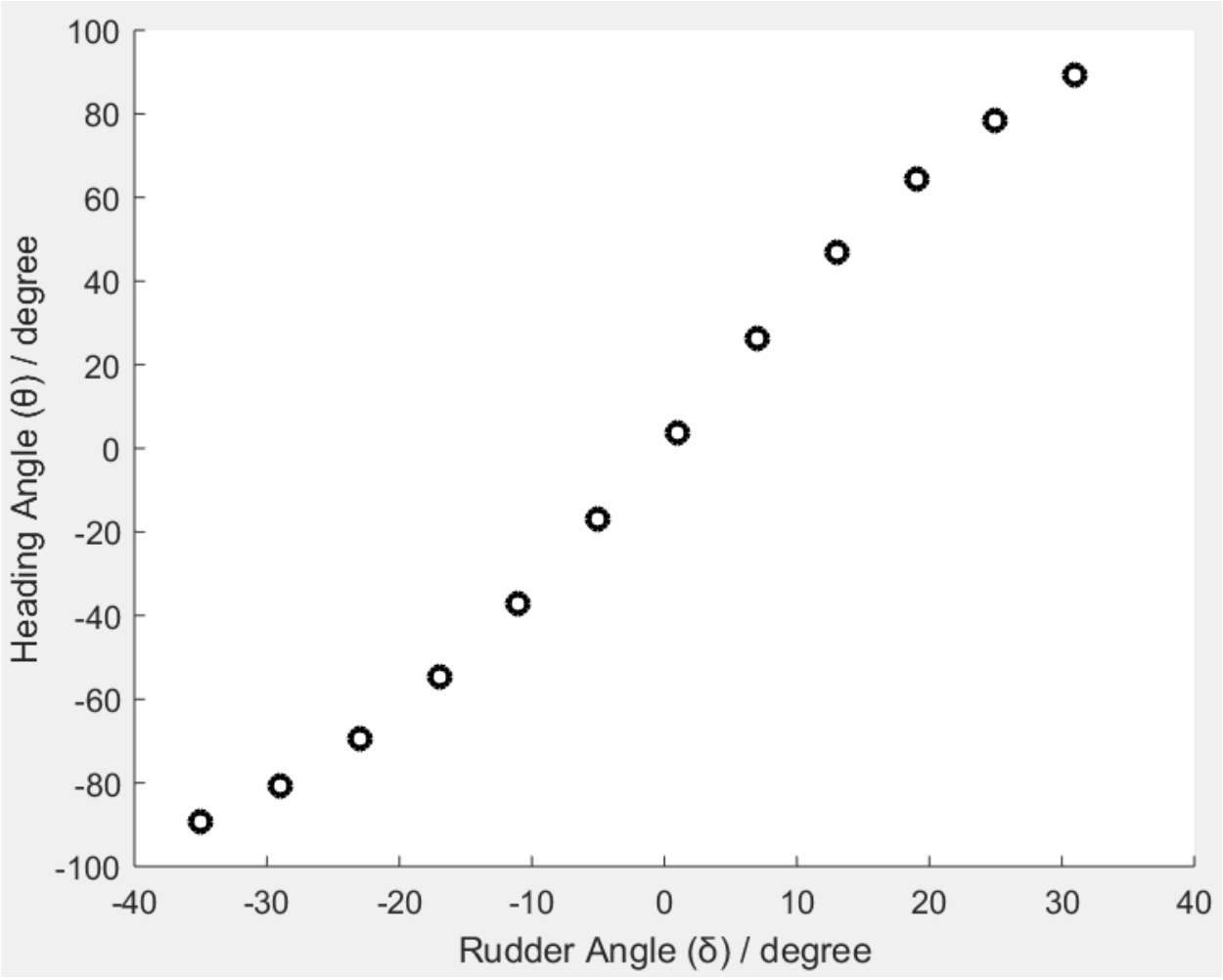

(a) (b)

**Figure 13.** The data of rudder angles and corresponding headings.

Thus, the correlation coefficient of the rudder angle and heading angle can be calculated:

$$r(\delta,\theta)=\frac{Cov(\delta,\theta)}{\sqrt{Var[\delta]\cdot Var[\theta]}}=0.9957$$

The result indicates that the rudder angle and heading angle indeed have a positive relation. Taking advantage of the function fitting method, the relational function can be expressed.

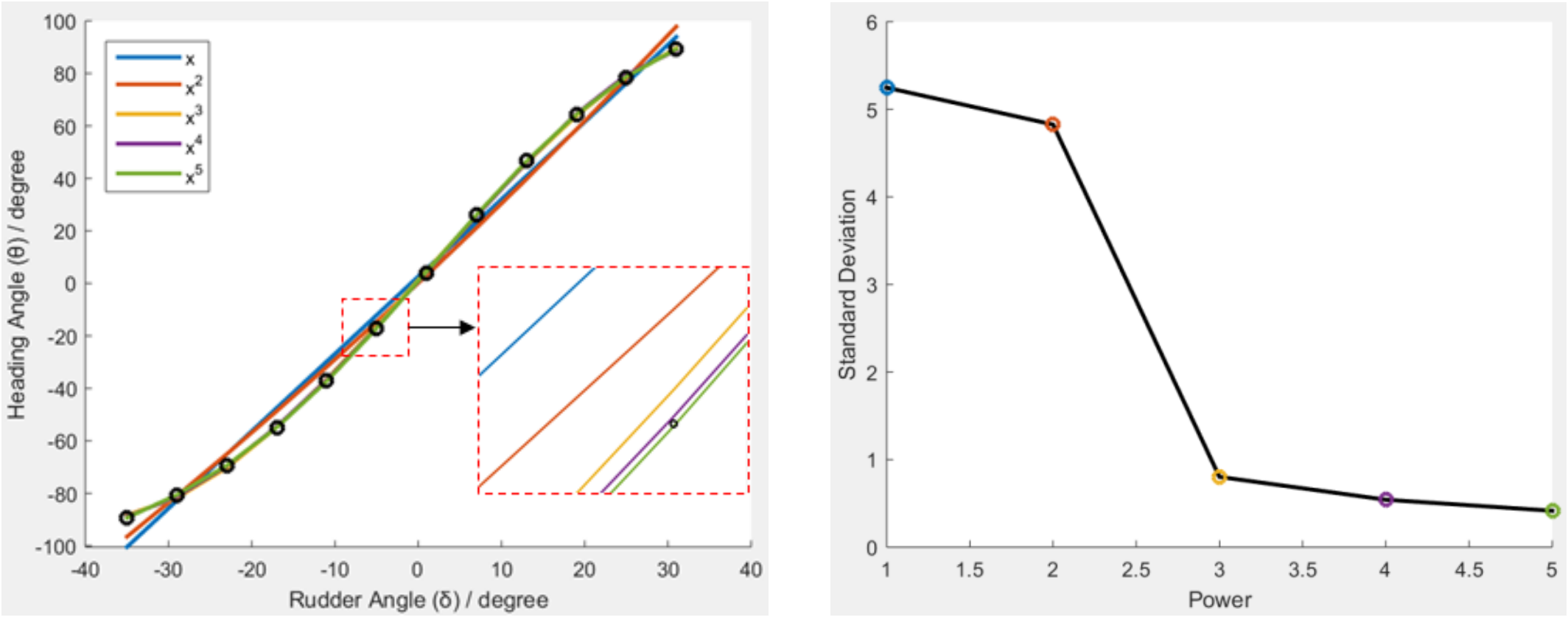

(a) (b)

**Figure 14.** The function fitting: (a) the results of different power; (b) the corresponding standard deviations.

Figure 14 (a) shows the results of different power function fitting and Figure 14 (b) shows the corresponding standard deviations. It can be seen that after third power function fittings, the change of standard deviation is small and the matching of the function tends to be stable. Thus, the cubic function is used to fit the relation of rudder angle and heading:

$$\theta = a\delta^3 + b\delta^2 + c\delta + d \tag{3}$$

where $a$, $b$, $c$, and $d$ are the fitting coefficient. Similarly, the rudder angle can also be expressed by heading ($\delta = f(\theta)$).

Apart from the relation between the rudder angle and heading, different rudder angle also results in different waypoints. So there is an indirect relation between rudder angle and position. To make this indirect relation more clear, it is necessary to transform the position data to another parameter.

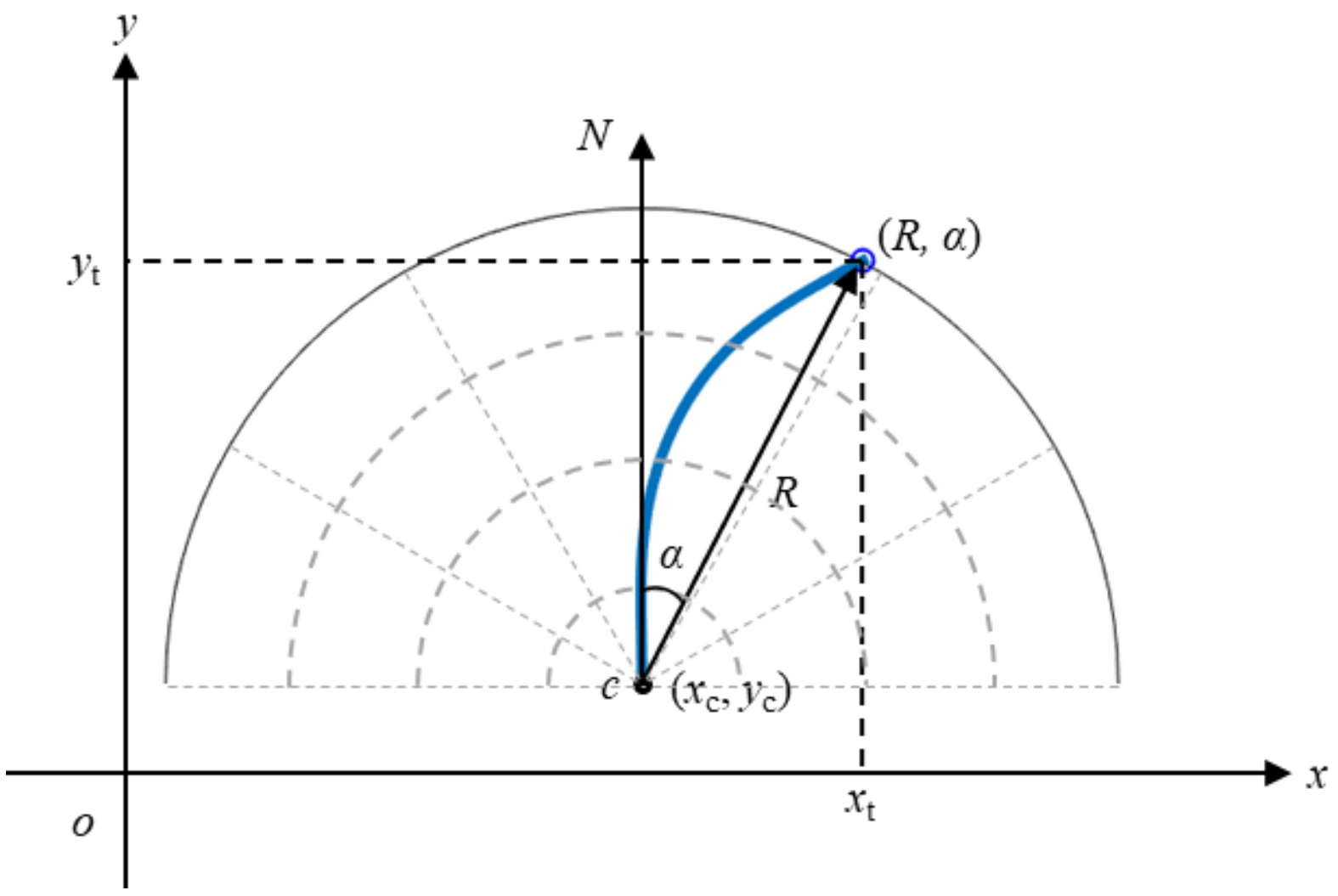

**Figure 15.** The Transformation between Polar coordinate system and the World coordinate system.

In Figure 15 it can be seen that the polar coordinate system is a usual way to deal with circle grid-related problems. Once the radius and the central angle are determined, the position can be expressed as ($R$, $\alpha$). Assuming that the coordinate of the center in the world coordinate system is ($x_c$, $y_c$), so the transformation from the Polar coordinate system to the World coordinate system can be expressed as:

$$\begin{cases} x_t = x_c + R \cdot \sin\alpha \\ y_t = x_c + R \cdot \cos\alpha \end{cases} \tag{4}$$

When $R$ is fixed, the position ($x_t$, $y_t$) is determined by the central angle $\alpha$.

Therefore, the central angle determines the position, the heading angle determines the direction and the rudder angle determines the manipulation. Taking advantage of the above function fitting method, the three angles can be transformed into each other (Figure 16). And the problem of solving the rudder angle can be replaced by the problem of solving the heading or the central angle.

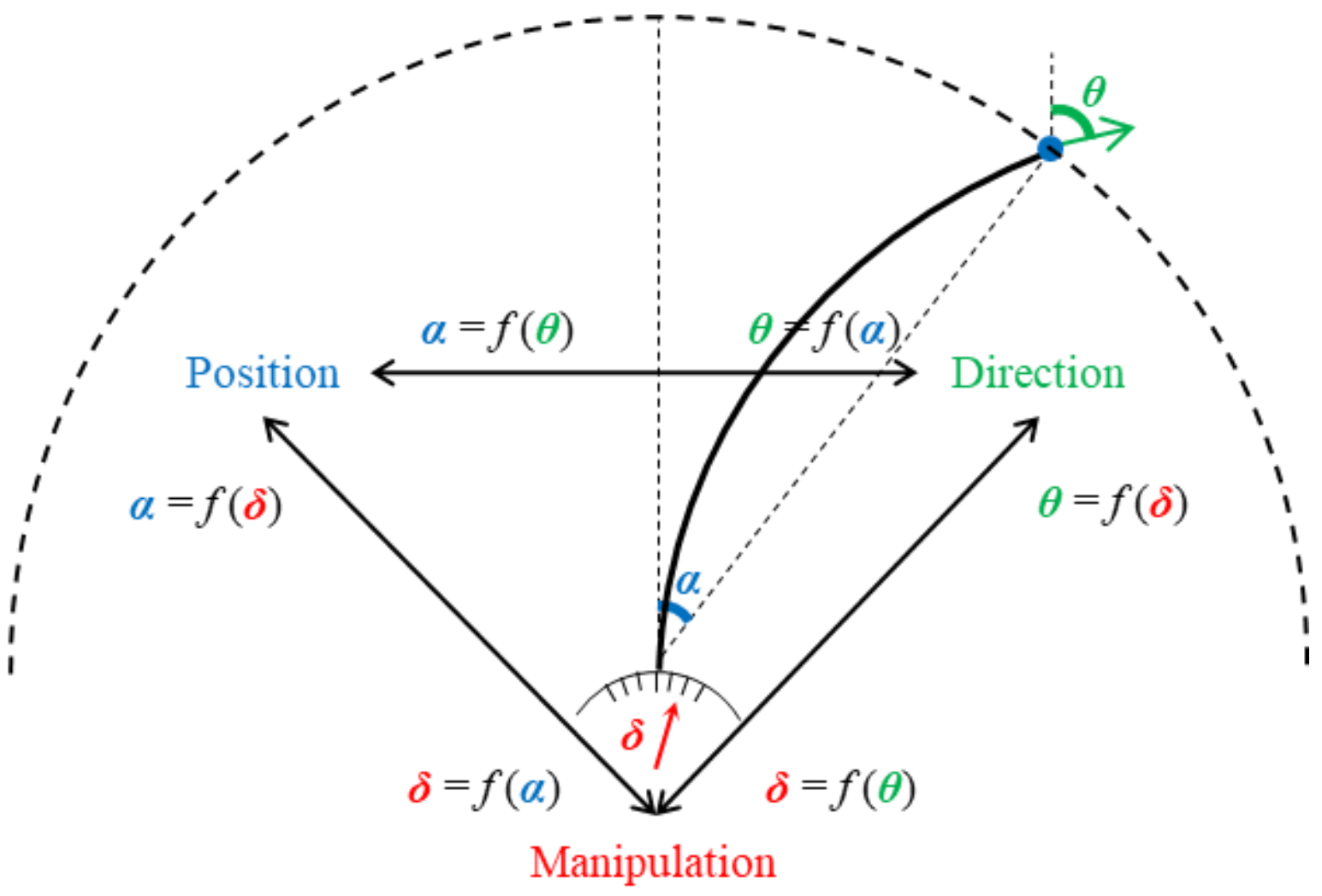

**Figure 16.** The Transformation among position ($\alpha$), direction ($\theta$), and Manipulation ($\delta$).

## 4.3 Heading Selecting and Calculating Algorithm

According to Section 4.2, the key to solving the rudder angle is to calculate the heading or central angle. The central angle determines the waypoint position. But in the circle grid, the position has no much direct relevance to the quality of the path. The heading angle determines the path direction. If the destination is determined, the optimal direction is toward it. Therefore, the heading angle can be used as an indicator to judge the path quality.

As is shown in Figure 17 (a), point *C* is the current node, point *D* is the destination and the circular sector shadow is the navigable area (trajectory cell set). Assuming point *T* is the next optimal node whose corresponding heading $\theta_t$ is toward point *D*, the angle can be calculated by the coordinate of *T* and *D*:

$$\theta_t = \arctan(\frac{x_d - x_t}{y_d - y_t}) \tag{5}$$

And the coordination of point *T* can be calculated by equation (4):

$$\begin{cases} x_t = x_c + R \cdot \sin\alpha \\ y_t = y_c + R \cdot \cos\alpha \end{cases} \tag{6}$$

Besides, by using the relational function of heading and central angle, $\theta_t$ can also be expressed as:

$$\theta_t = f(\alpha) \tag{7}$$

Combining equation (5), (6), and (7) can observe the only one variate equation:

$$f(\alpha) = \arctan(\frac{x_d - (x_c + R \cdot \sin\alpha)}{y_d - (y_c + R \cdot \cos\alpha)}) \tag{8}$$

Thus, the central angle $\alpha$ can be determined, and the other two angles are also solved.

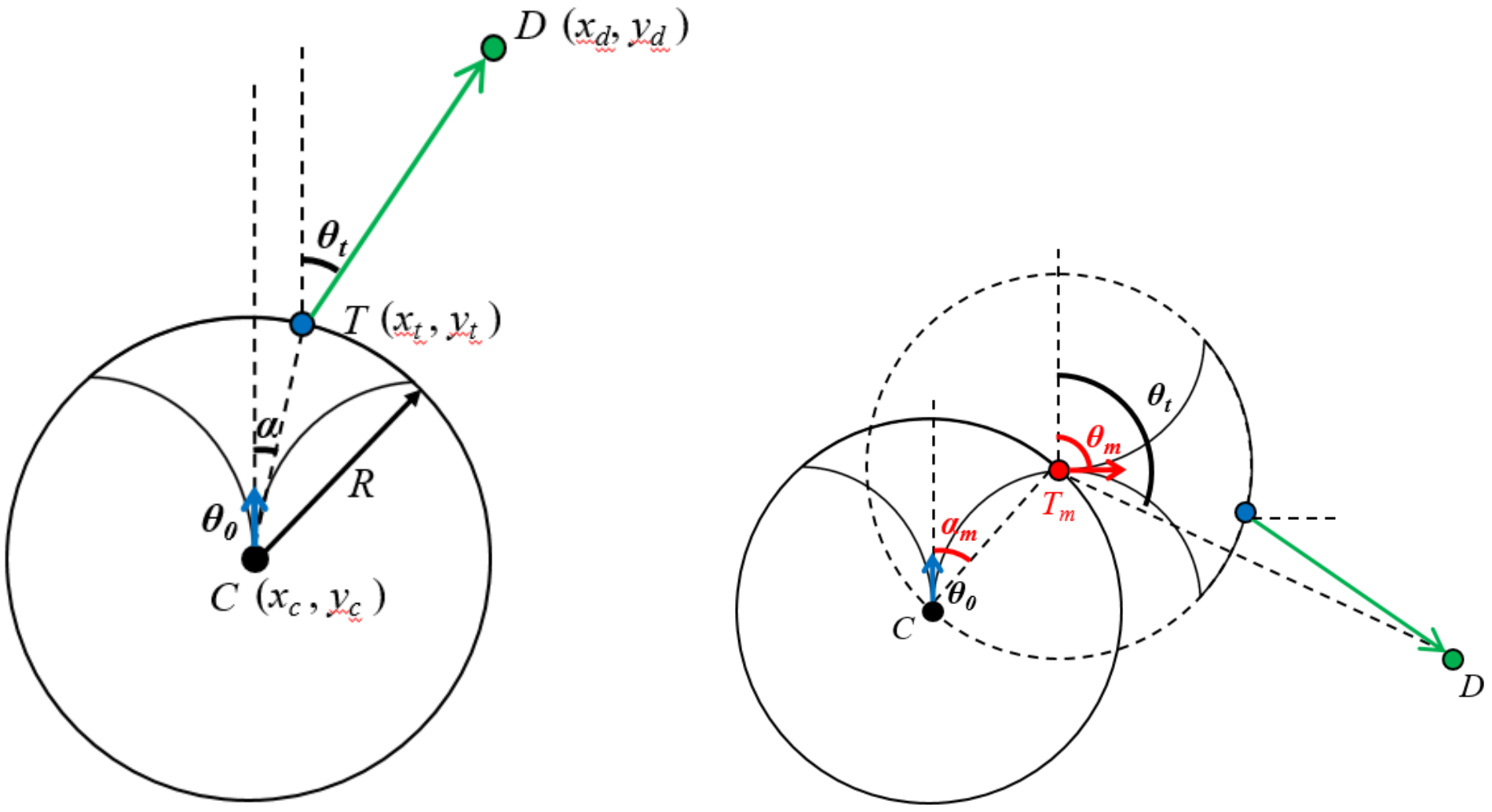

(a) (b)

**Figure 17.** The strategy for searching: (a) the acute angle situation; (b) the obtuse angle situation.

However, in some situations, the heading angle cannot be reached directly in one step because of the restriction of the trajectory cell. As is shown in Figure 17 (b), when the destination is out of the range of the trajectory cell covering, the $\theta_t$ is over the maximum heading angle $\theta_m$ ($\theta_t$ > $\theta_m$). Thus, firstly the USV has to steer a maximum rudder angle to get the edge ($T_m$) of the navigable area with the maximum heading change. Then, takes the edge point as the current point to execute the above searching step.

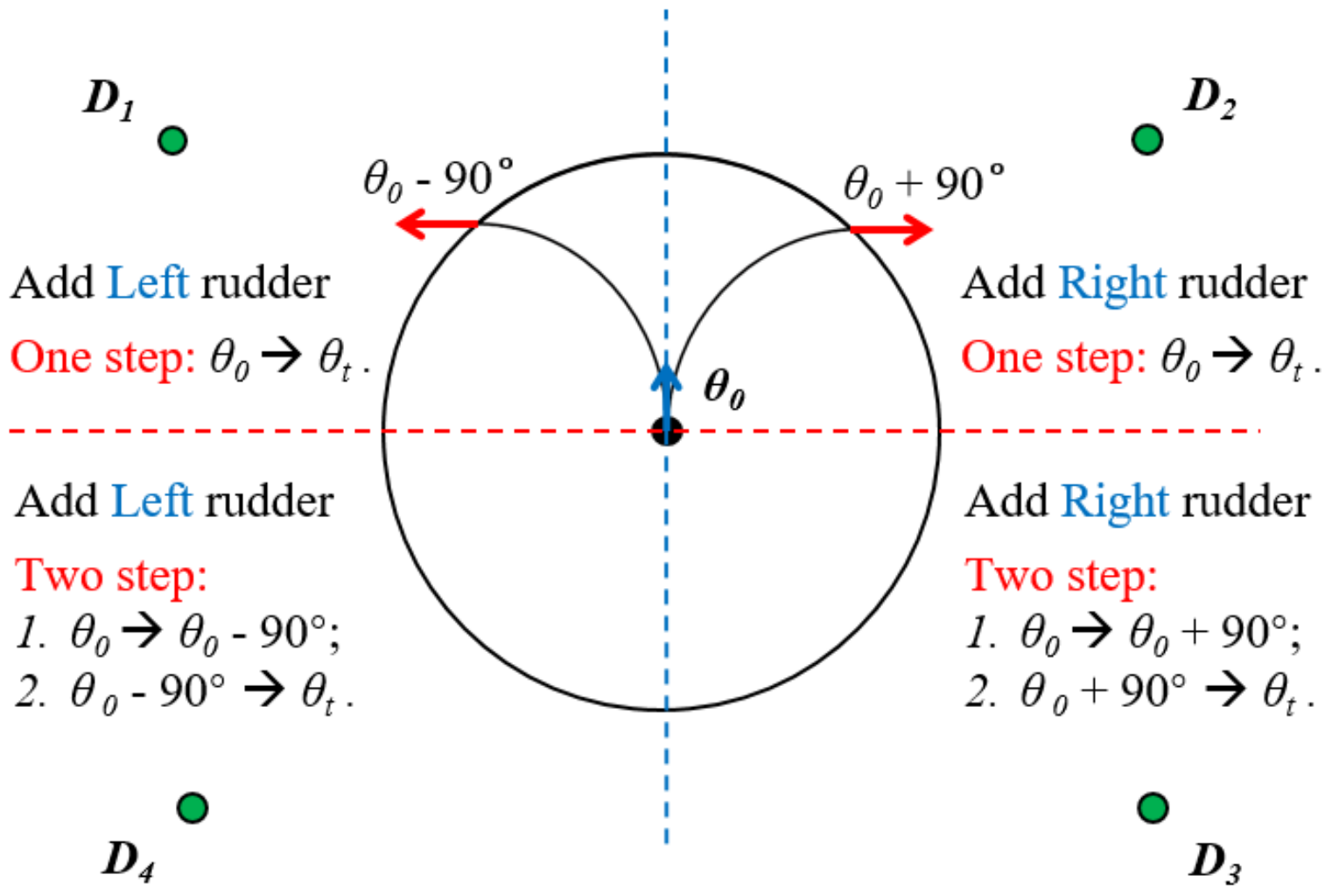

**Figure 18.** The searching strategy of four different destinations.

In fact, according to Section 3.5, the maximum heading change of the trajectory cell is 90°. So there are four different situations for the destination position. As is shown in Figure 18, when it locates over the red line ($D_1$ and $D_2$), there we only need one steering to change the heading toward the destination. When it below the red line ($D_3$ and $D_4$), there will be two steps. Firstly, add a bigger rudder angle to make the heading change 90°, so the angle between heading and destination is acute. Then, we can select the optimal heading to make it toward the destination. Thus, before searching, the relative location between the current node and destination should be judged first.

## 4.4 Strategy for Static Obstacle Avoidance

In waters with obstacles, collision avoidance becomes the primary task. There are plenty of methods to deal with this problem. Considering the characteristic of the circle grid and the important role of the heading in the process of path searching, this paper uses a circle to cover the obstacle and find the best tangency point to bypass it. So the main idea is to compare the angles between the destination and the tangency points.

However, if the heading angle is determined by the way of Section 4.3, the calculating will be complex. Because different headings correspond to different waypoints, the angles made up of radial are not based on the same point, and they cannot be compared directly. As is shown in Figure 19 (a), the four different lines are from different points on the circle arc. To simplify the calculation procedure, the paper takes the current point (central point) as the basic point.

It can be seen in Figure 19 (b), $N$ points to the true north, the position of the origination $C$ ($x_C$, $y_C$), destination $D$ ($x_D$, $y_D$), and obstacle O ($x_O$, $y_O$) are known. And the radius of the circle grid and the obstacle are $R$ and $r$ respectively. So the tangency point $A_1$ and $B_1$ in the current state can be determined by the angle $\angle NCA_1$ and $\angle NCB_1$ respectively:

$$\angle NCA_1 = \angle NCO\text{-}\angle A_1CO$$
$$\angle NCB_1 = \angle NCO + \angle B_1CO$$

The $\angle NCO$ can be calculated by node $C$ and $O$:

$$\angle NCO = \arctan\left(\frac{x_O - x_C}{y_O - y_C}\right)$$

And the $\angle A_1CO$ is equal to $\angle B_1CO$, which can be calculated by node $O$, $C$, and $r$:

$$\angle A_1CO = \angle B_1CO = \arcsin\left(\frac{r}{\sqrt{(x_O - x_C)^2 + (y_O - y_C)^2}}\right)$$

Thus, $\angle NCA_1$ and $\angle NCB_1$ can be expressed as:

$$\angle NCA_1 = \arctan\left(\frac{x_O - x_C}{y_O - y_C}\right) - \arcsin\left(\frac{r}{\sqrt{(x_O - x_C)^2 + (y_O - y_C)^2}}\right)$$

$$\angle NCB_1 = \arctan\left(\frac{x_O - x_C}{y_O - y_C}\right) + \arcsin\left(\frac{r}{\sqrt{(x_O - x_C)^2 + (y_O - y_C)^2}}\right)$$

The best heading is the one that has the smaller difference between the two tangency angles and the destination angle ($\angle NCD$). It is obviously in this example that $\angle A_1CD$ is larger than $\angle B_1CD$. So $\angle NCB_1$ is the heading of the next node, and this node ($T_1$) can also be determined by the relational function.

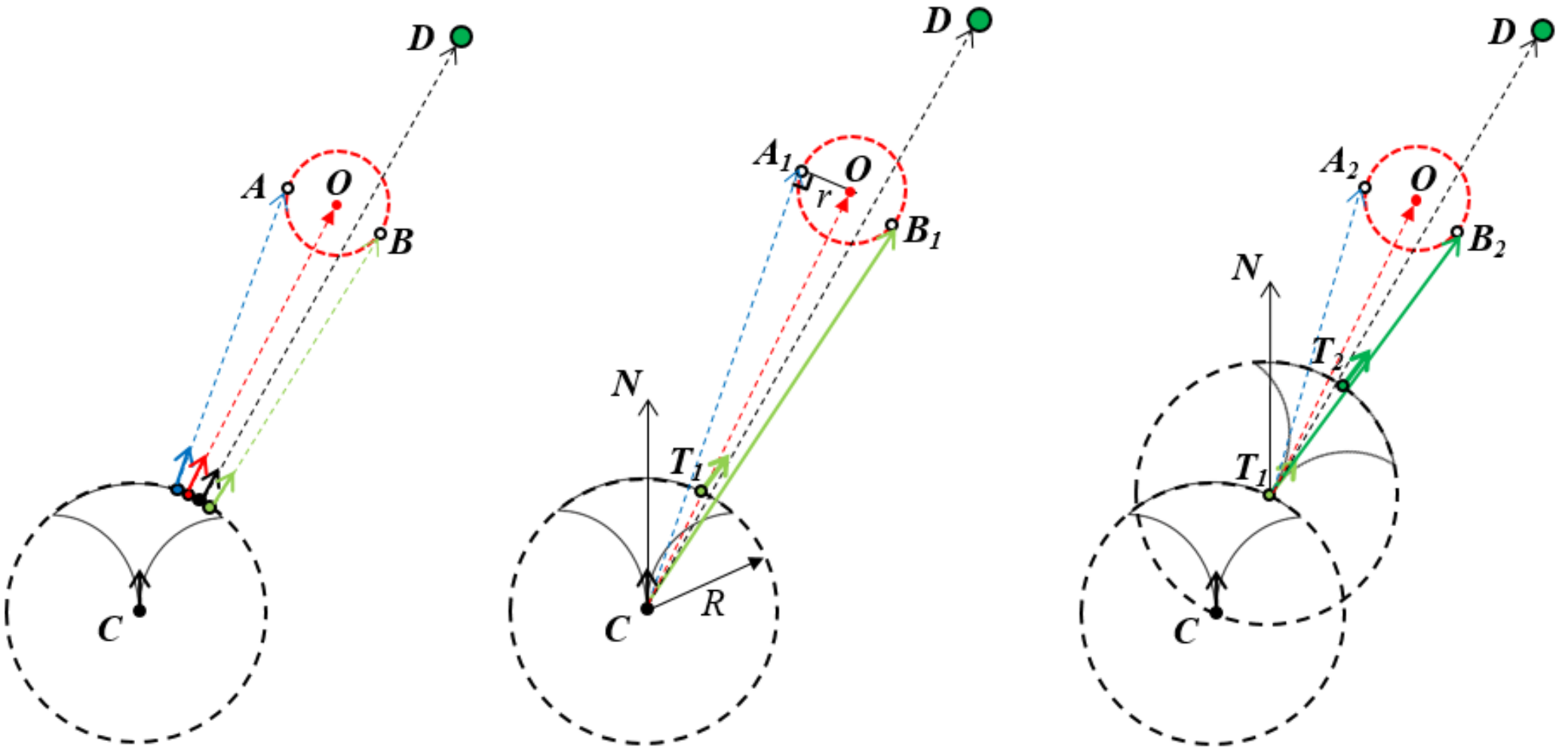

(a) (b) (c)

**Figure 19.** The strategy for obstacle avoidance: (a) the heading determining way in free waters; (b) the simple way in obstacle waters; (c) the reducing error way of continuous tracking obstacle.

However, because of the simplification, node $T_1$ is not on the line of $CB_1$. If the heading is fixed in this direction, there is still a possibility for the USV to crash on the obstacle. Thus, the above process should be continued. It can be seen in Figure 19 (c), after determining the next node, the $T_1$ becomes the current node. Based on this node, the new tangency points ($A_2$ and $B_2$) can be calculated and the new heading (on node $T_2$) can be determined. It is obvious that the distance from $T_2$ to $T_1B_2$ is shorter than the distance from $T_1$ to $CB_1$. That means the error of the simplification can be reduced by continuous tracking obstacles.

When it comes to multi obstacles, the searching process is more complex. Firstly, the tangency points of each obstacle circle should be calculated to find the minimum and maximum tangency angles. Then, compared to the destination angle, select the tangency point with a smaller angle difference. So the circle with the selected tangency point is the currently focused obstacle. After the USV bypasses it, this obstacle is neglected, and the algorithm recalculates the best tangency points in the rest obstacles. Finally, until there are no obstacles that need to be concerned, the heading is assigned to point at the destination.

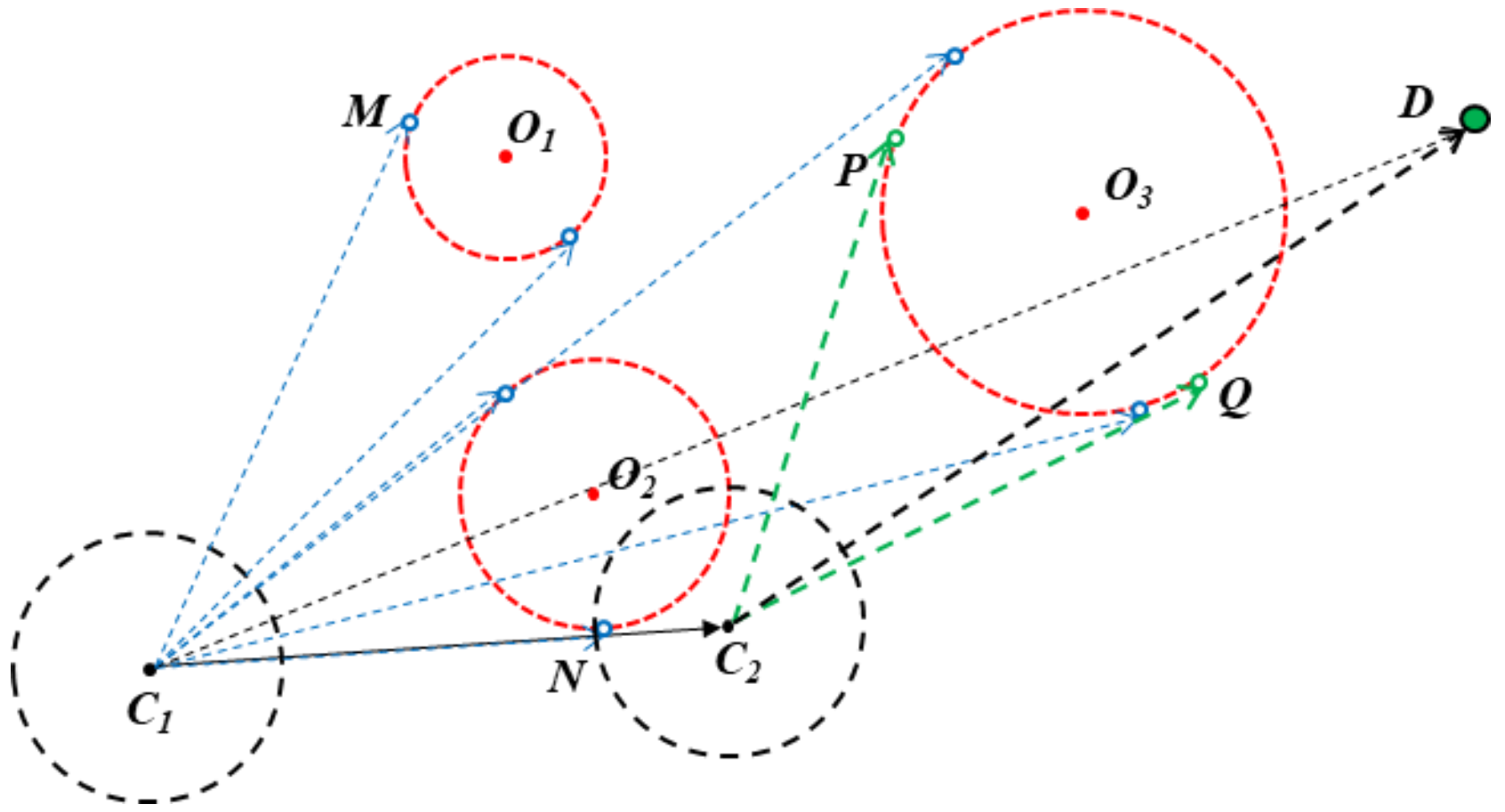

**Figure 20.** The strategy for multi obstacle collision avoidance.

As the example shown in Figure 20, there are three obstacles in the waters. At the beginning, six tangency points are calculated based on the original point ($C_1$), and node $M$ and $N$ correspond to the minimum and maximum tangency angle. Because $\angle MC_1D$ is larger than $\angle NC_1D$, node $N$ is selected. After the USV reaches point $C_2$, Obstacle $O_2$ is bypassed and there is no need to consider it (neglect). The algorithm starts to focus on Obstacle $O_3$ (Obstacle $O_1$ is also be neglected). So the tangency points $P$ and $Q$ are the next points waiting to be compared. Namely, the next searching process is to deal with a single obstacle.

## 4.5 Strategy for Dynamic Obstacle Avoidance

When it comes to the moving obstacle, there will be two situations for the USV: maintain the heading or have a steering command. So before taking the actions, the judgment should be made.

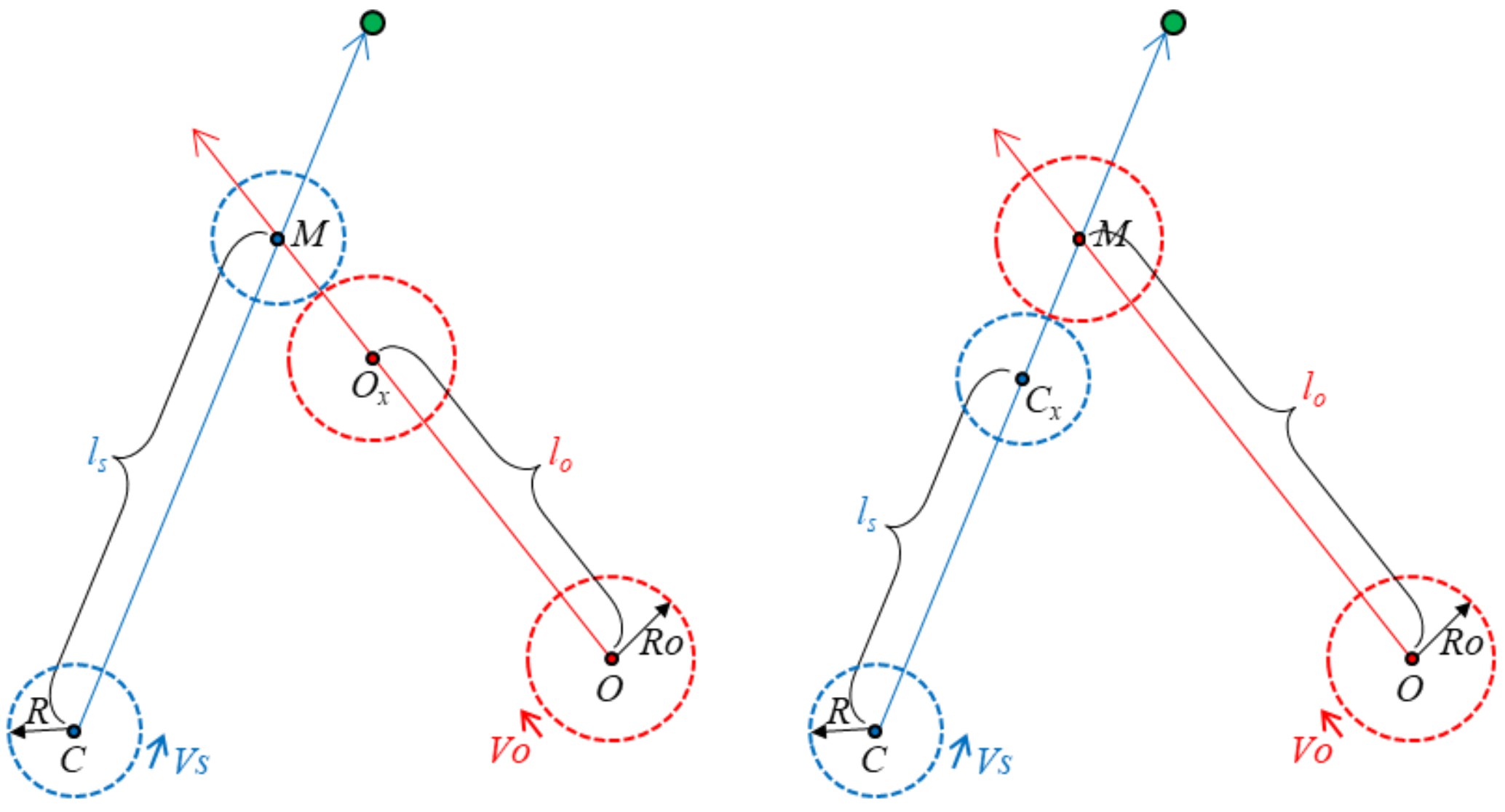

(a) (b)

**Figure 21.** The situations that the USV maintains the heading.

Assuming the obstacle maintains its course and speed, the USV maintains its speed (stable velocity) but can change the heading. If the USV doesn't have to change its heading, $M$ is the intersection point from the heading extending the line of USV and obstacle. When the USV reaches point $M$ first (show in Figure 21 (a)), to avoid the collision, the critical situation is that the obstacle is located at position $O_x$ (circle $O_x$ is tangent to circle $M$). Thus, according to the relationship between motion and geometry at the moment:

$$t = \frac{l_s}{V_s} = \frac{\sqrt{(x_m - x_c)^2 + (y_m - y_c)^2}}{V_s} \tag{9}$$

$$l_o = t \cdot V_O \tag{10}$$

$$l_o = \leq OM - (R + R_o) = \sqrt{(x_m - x_o)^2 + (y_m - y_o)^2} - (R + R_o) \tag{11}$$

Combining equation (9), (10), and (11), we can observe the condition that the speed of the obstacle ($V_O$) has to comply:

$$V_O \leq \frac{\sqrt{(x_m - x_o)^2 + (y_m - y_o)^2} - (R + R_o)}{\sqrt{(x_m - x_c)^2 + (y_m - y_c)^2}} \cdot V_S \tag{12}$$

where $x_m$, $y_m$ are the coordinates of node $M$, $x_c$, $y_c$ are the coordinates of node $C$, $x_o$, $y_o$ are the coordinates of node $O$, $Vs$ is the velocity of the USV, $R$, and $R_0$ are the radius of circle cell and obstacle respectively.

Similarly, when the obstacle reaches point $M$ first (show in Figure 21 (b)), to avoid the collision, the critical situation is that the USV is located at position $C_x$ (circle $C_x$ is tangent to circle $M$). So $V_O$ has to comply:

$$V_O \geq \frac{\sqrt{(x_m - x_o)^2 + (y_m - y_o)^2}}{\sqrt{(x_m - x_c)^2 + (y_m - y_c)^2} - (R + R_o)} \cdot V_S \tag{13}$$

Thus, if $V_O$ conforms to equations (12) and (13), the USV can maintain its heading. However, when $V_O$ is between the two values, the USV has to have a steering command to avoid the collision. Based on the circle grid, this paper proposes a method to transforms the dynamic obstacle into a static one.

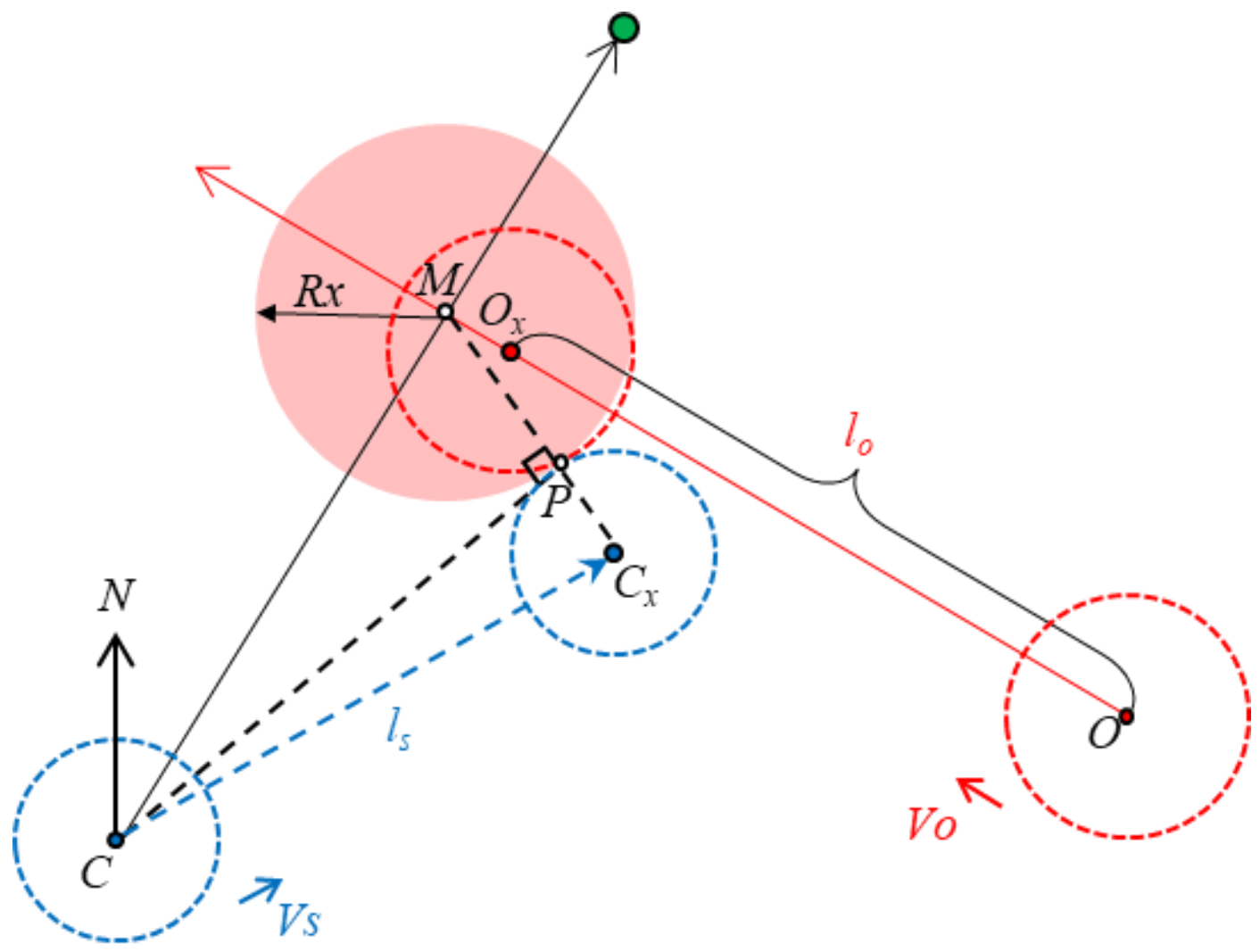

**Figure 22.** The situation that the USV has to have a steering command.

It can be seen in Figure 22, taking node $M$ as the central and $Rx$ as the radius virtualizes a static obstacle (the red disc). When the dynamic obstacle reaches to node $O_x$, the USV is located at node $C_x$ and the circle $C_x$ is tangent to circle $M$ at point $P$. To avoid the collision, the distance between $O_x$ and $C_x$ has to comply:

$$O_xC_x=\sqrt{(x_{ox}-x_{cx})^2+(y_{ox}-y_{cx})^2} \geq R+R_o \tag{14}$$

The coordinates of node $C_x$ can be calculated by $CC_x$ ($l_s$) and $\angle NCC_x$:

$$\begin{cases} x_{cx}=x_c+l_s\cdot\sin\angle NCC_x \\ y_{cx}=y_c+l_s\cdot\cos\angle NCC_x \end{cases} \tag{15}$$

$l_s$ and $\angle NCC_x$ can be determined by:

$$\begin{aligned} l_s &= \sqrt{CP^2+PC_x^{\,2}} = \sqrt{CM^2-MP^2+PC_x^{\,2}} \\ &= \sqrt{(x_c-x_m)^2+(y_c-y_m)^2\text{-}R_x^{\,2}+R^2} \end{aligned} \tag{16}$$

$$\begin{aligned} \angle NCC_x &= \angle NCM+\angle MCP+\angle PCC_x \\ &= \arctan(\frac{x_m-x_c}{y_m-y_c})+\arcsin(\frac{R_x}{\sqrt{(x_c-x_m)^2+(y_c-y_m)^2}}) \\ &\quad +\arcsin(\frac{R}{\sqrt{(x_c-x_m)^2+(y_c-y_m)^2\text{-}R_x^{\,2}+R^2}}) \end{aligned} \tag{17}$$

The coordinates of node $O_x$ can be calculated by $OO_x$ ($l_0$) and the course of the obstacle ($\theta_o$):

$$\begin{cases} x_{ox}=x_o+l_o\cdot\sin\theta_o \\ y_{ox}=y_o+l_o\cdot\cos\theta_o \end{cases} \tag{18}$$

$l_o$ can be determined by the time of the motion:

$$l_o=V_O\cdot t \tag{19}$$

$$t = \frac{l_s}{V_S} = \frac{\sqrt{(x_c - x_m)^2 + (y_c - y_m)^2 \text{-} R_x{}^2 + R^2}}{V_S} \tag{20}$$

Combining equation (14) - (20), we can have the minimum value of $R_x$.

# 5. Experiments and Results

This section provides some experiments to show the results of the method. Section 5.1 is a comparative experiment. The experiment firstly compares the path characteristic of grid-based methods and circle grid-based methods. Then, the experiment of static obstacle avoidance comparison is to illustrate the differences of the two kinds of methods dealing with static obstacles. Section 5.2 analyzes all three situations in dynamic obstacle avoidance. The experiment demonstrates some details in the process of obstacle avoidance and the validity of the proposed method. In the experiments, the parameters of a real ship are used to carry out the experiments, where the ship length is 63.6 m, the width is 16.4 m, the draft is 6.22 m, the rudder radius is 3.6 m, the tonnage is 4522 ton, the minimum rudder angle is -35°, the maximum rudder is 35°, the propeller speed per second is 180 rpm. All the simulation experiments are carried out by MATLAB 2018a.

## 5.1 Comparative Experiments

(1) Path Characteristic Comparative

This experiment is to compare the path characteristics. It can be seen in Figure 23, (a) is the path without considering USV dynamic constraints, (b) is the motion path based on a grid, (c) is the motion path based on a circle grid.

Path (a) is made up of line segments without considering dynamic constraints. Path (b) and (c) are curves, but the characteristics are different. Based on a grid map, each step in (b) is on the node. But as the grid restraint, there are only three initial headings (0 °, 45 °, and 90 °) from 0 ° to 90 °. Path (c) has no direction restraint and any initial headings from 0 ° to 90 ° can be planned. It can also be seen that compared to path (b), path (c) is more efficient: the distance is shorter and the steering is less.

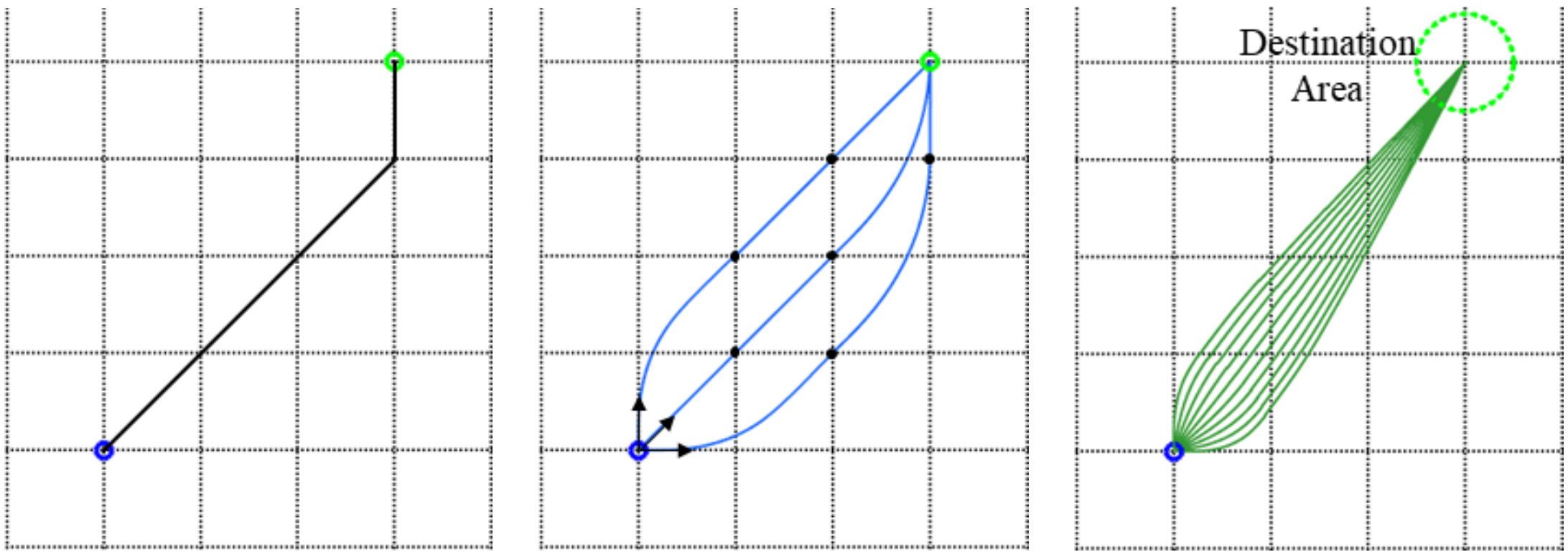

(a) (b) (c)

**Figure 23.** The paths with different algorithms: (a) the path without considering USV dynamic constraints; (b) the motion path based on a grid; (c) the motion path based on a circle grid.

In addition, the steering command of each step can be known. Table 1 is the series command rudder angles for ten different initial headings. It can be seen that because the angle between origination and destination is 37°, the first command rudder angle is from the positive value (right rudder) decreasing to a negative value (left rudder). And the reason why all the last command rudder angles are not 0° is that the rudder effect on the starboard side is better than the port side (mentioned in Section 3.5).

**Table 1.** The command rudder angle with different initial headings

| Initial Heading | Command Rudder Angle | | | | |
|---|---|---|---|---|---|
| 0° | 9.7347° | 1.2782° | 0.1592° | -0.0594° | -0.1460° |
| 10° | 6.8898° | 0.9384° | 0.0912° | -0.0798° | -0.1602° |
| 20° | 4.2221° | 0.5614° | 0.0126° | -0.1045° | -0.1776° |
| 30° | 1.6495° | 0.1592° | -0.0732° | -0.1321° | -0.1966° |
| 40° | -0.9101° | -0.2532° | -0.1619° | -0.1607° | -0.2153° |
| 50° | -3.5389° | -0.6599° | -0.2484° | -0.1880° | -0.2311° |
| 60° | -6.3190° | -1.0462° | -0.3283° | -0.2120° | -0.2425° |
| 70° | -9.3325° | -1.4000° | -0.3984° | -0.2316° | -0.2490° |
| 80° | -12.6616° | -1.7136° | -0.4570° | -0.2463° | -0.2510° |
| 90° | -16.3884° | -1.9839° | -0.5042° | -0.2565° | -0.2497° |

Thus, the circle grid-based algorithm not only satisfies the USV dynamic constraints but also can plan out a more efficient path and provide the steering command.

(2) Static Obstacle Avoidance Comparative

The avoidance becomes the priority for the USV in the obstacle space. In this experiment, the algorithm of grid-based and circle-based are compared.

The first experiment deals with one obstacle. The results are shown in Figure 24. The obstacle in (a) and (b) is small, the grid-based motion path (path (a)) has three steerings while there is only one steering in circle grid-based path (path (b)). The obstacle in (c) and (d) is big and the differences are more obvious. To avoid the obstacle, path (c) has two large steering commands at the beginning, which costs too much distance for the path. The path in (d) is gentler in that it has the same curve trend with the obstacle when bypassing it. Thus, when dealing with one single obstacle, the circle grid-based algorithm has a better motion path.

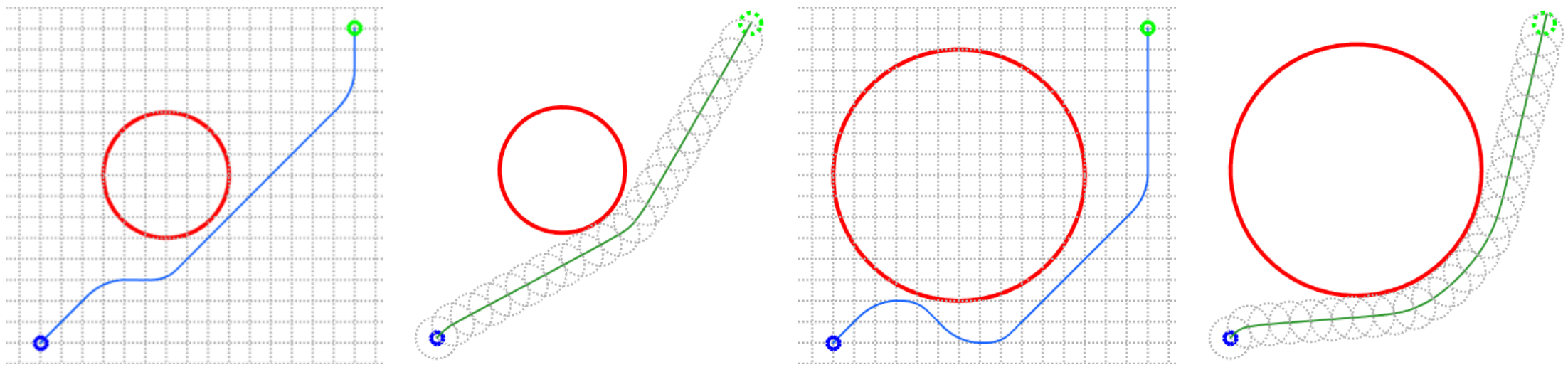

(a) (b) (c) (d)

**Figure 24.** The paths in one obstacle environment: (a) and (b) are the motion path based on grid and circle grid respectively with a small obstacle; (c) and (d) are the motion paths based on grid and circle grid respectively with a big obstacle.

The second experiment deals with multi obstacles. The scenario in this experiment is more complex. It can be seen in Figure 25, the two paths choose a similar way to get to the destination. The distance cost and the times of steering are shown in Table 2. The distance cost in circle grid-based algorithm is a little lower than grid-based. But for the times of steering, the former is half of the latter. Thus, when dealing with multi obstacles, the circle grid-based algorithm is also better than grid-based one.

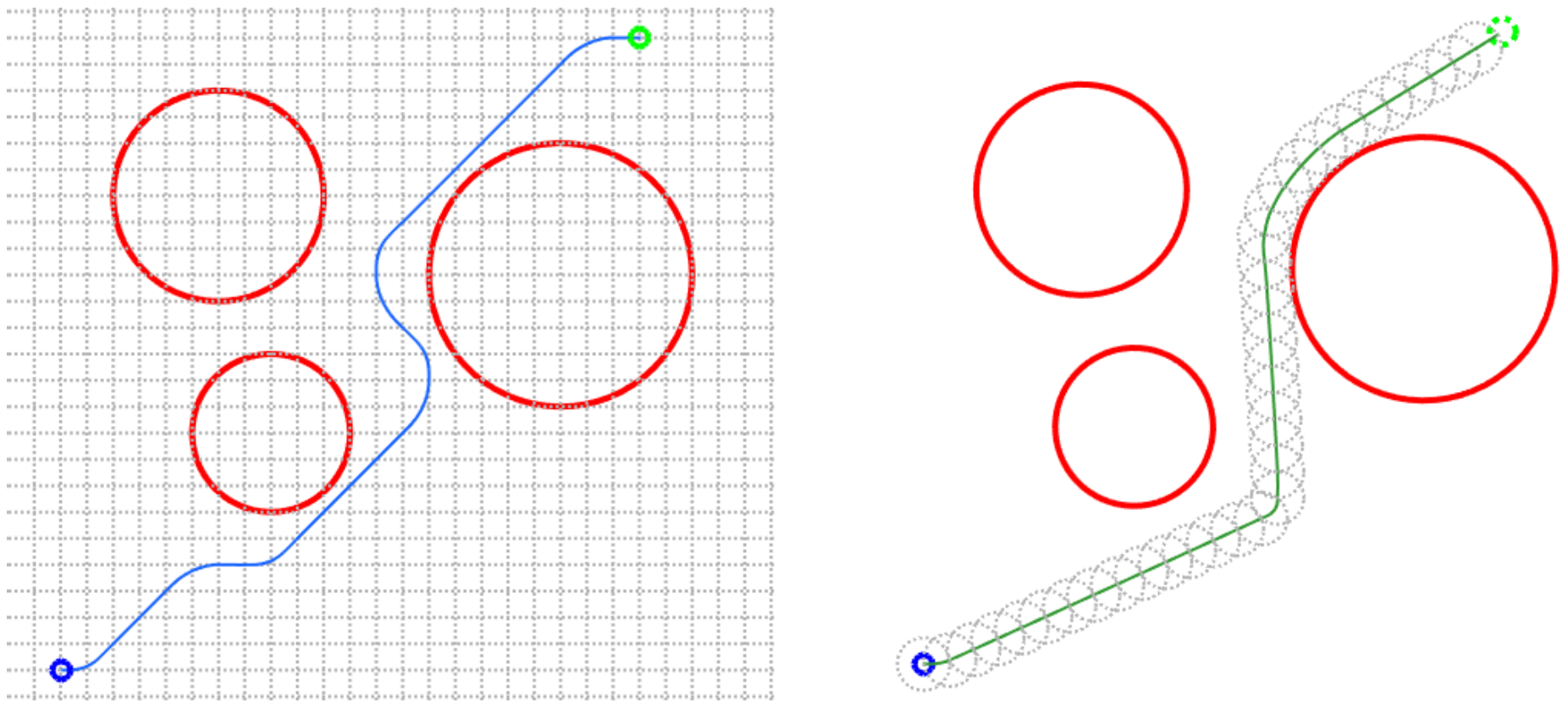

(a) (b)

**Figure 25.** The paths in multi obstacles environment: (a) is the motion path based on grid; (b) is the motion path based on circle grid.

Table 2. The distance cost and the times of steering for the two algorithms

| Algorithm | Grid-Based | Circle-Based |
|---|---|---|
| Distance (units) | 40 | 37 |
| Steering (times) | 6 | 3 |

## 5.2 Dynamic Obstacle Avoidance Experiments

Three situations are carried out in this simulation experiment. Different ship domains (circle) are set for the own ship (USV) and the other ship (moving obstacle): the domain radius of the own ship is 600 m, the othe ship is 900 m.

In the first situation, the velocities of the USV and the obstacle conform to equation (17), the experiment process is shown in Figure 26. Because the USV is faster than the obstacle, when the USV arrives at the intersection point, the obstacle is still far from it (Figure 26 (b)). The USV doesn't have to steer. The changes of distance between the USV and the obstacle are shown in Figure 27. It can be seen that the minimum distance is 1808 m which is larger than the sum of the two ship domain radius (1500 m).

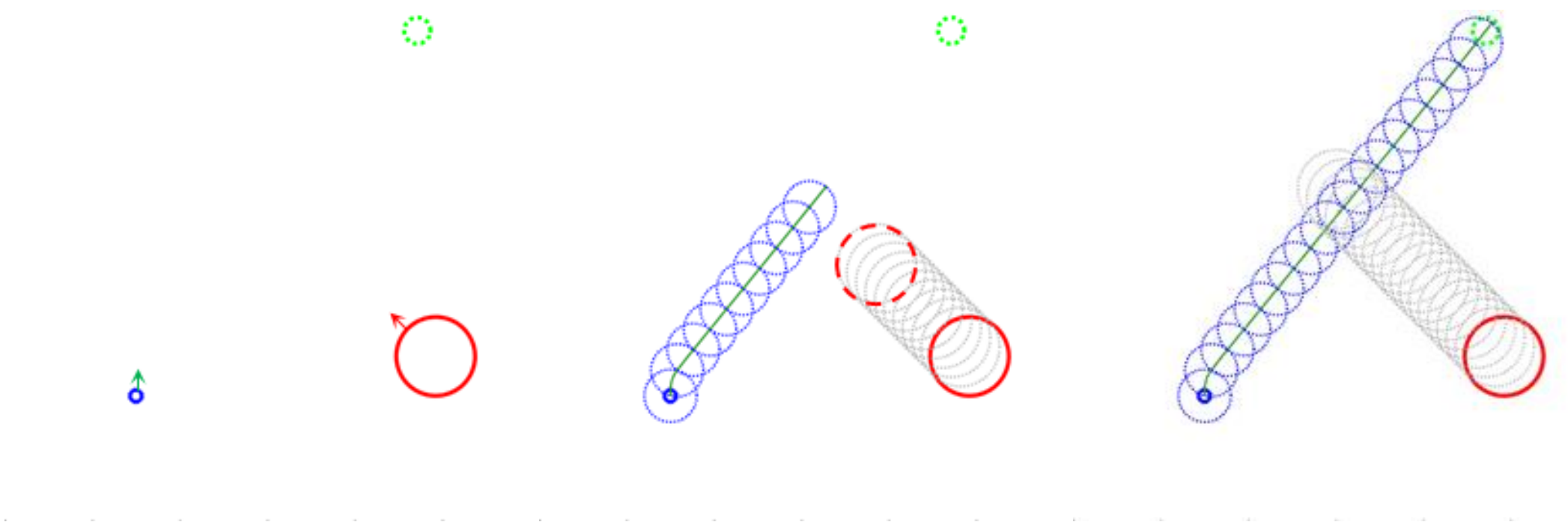

(a) (b) (c)

**Figure 26.** The process of dynamic obstacle avoidance (conform to equation (17)): (a) is the initial state; (b) is the critical state; (c) is the final state.

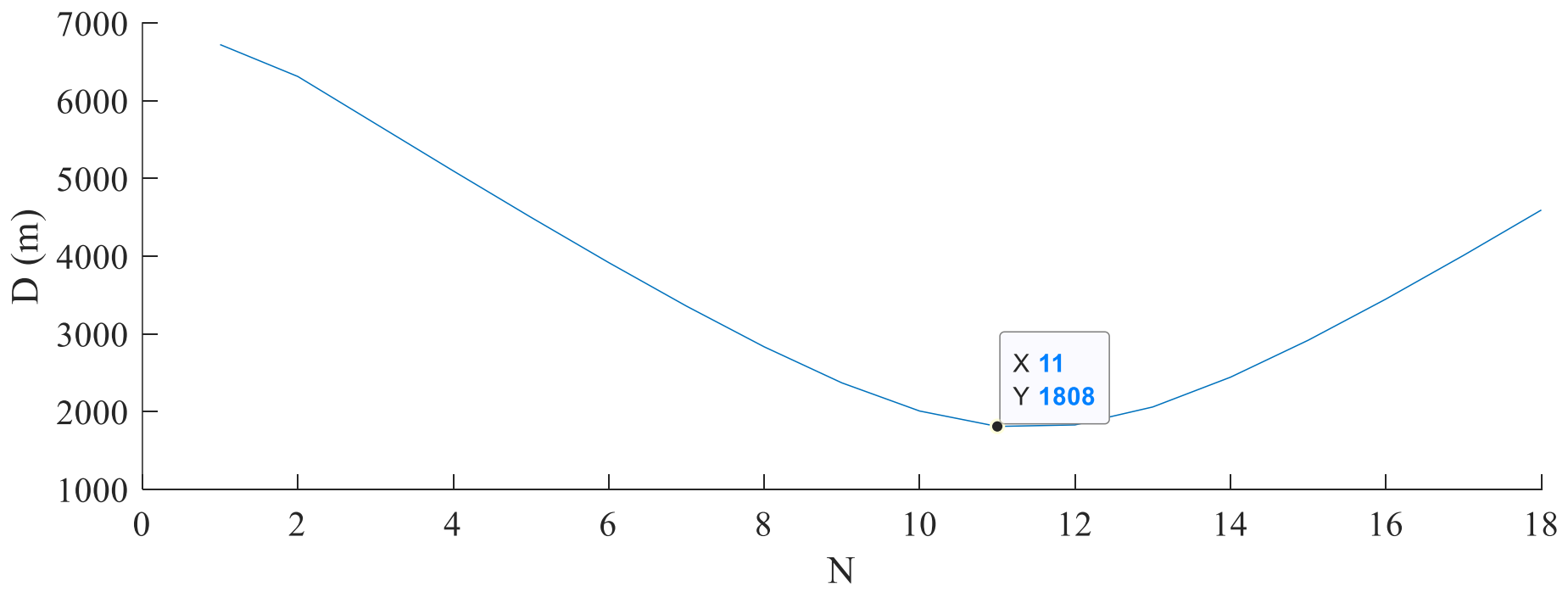

**Figure 27.** Distance between the USV and the obstacle in situation 1.

In the first situation, the velocities of the USV and the obstacle conform to equation (13), the experiment process is shown in Figure 28. Because the obstacle is faster than the USV, when the obstacle arrives at the intersection point, the USV is still far from it (Figure 28 (b)). The USV also doesn't have to steer. The changes of distance between the USV and the obstacle are shown in Figure 29. It can be seen that the minimum distance is 1937 m which is also larger than the sum of the two ship domain radius.

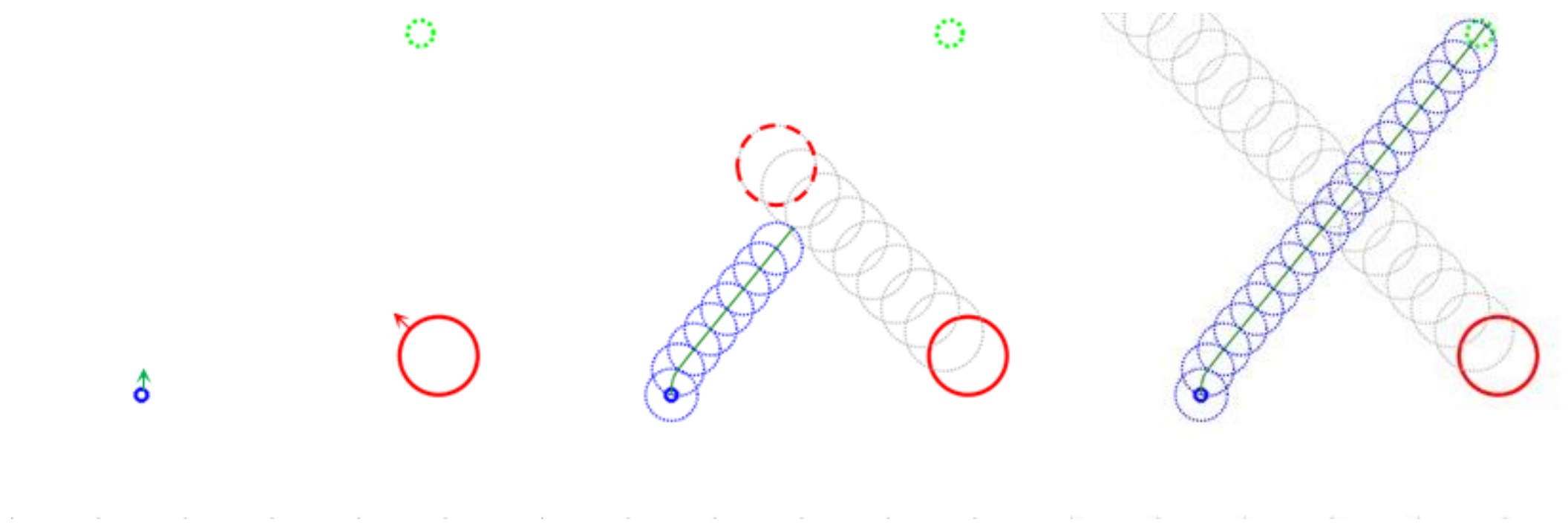

(a) (b) (c)

**Figure 28.** The process of dynamic obstacle avoidance (conform to equation (13)): (a) is the initial state; (b) is the critical state; (c) is the final state.

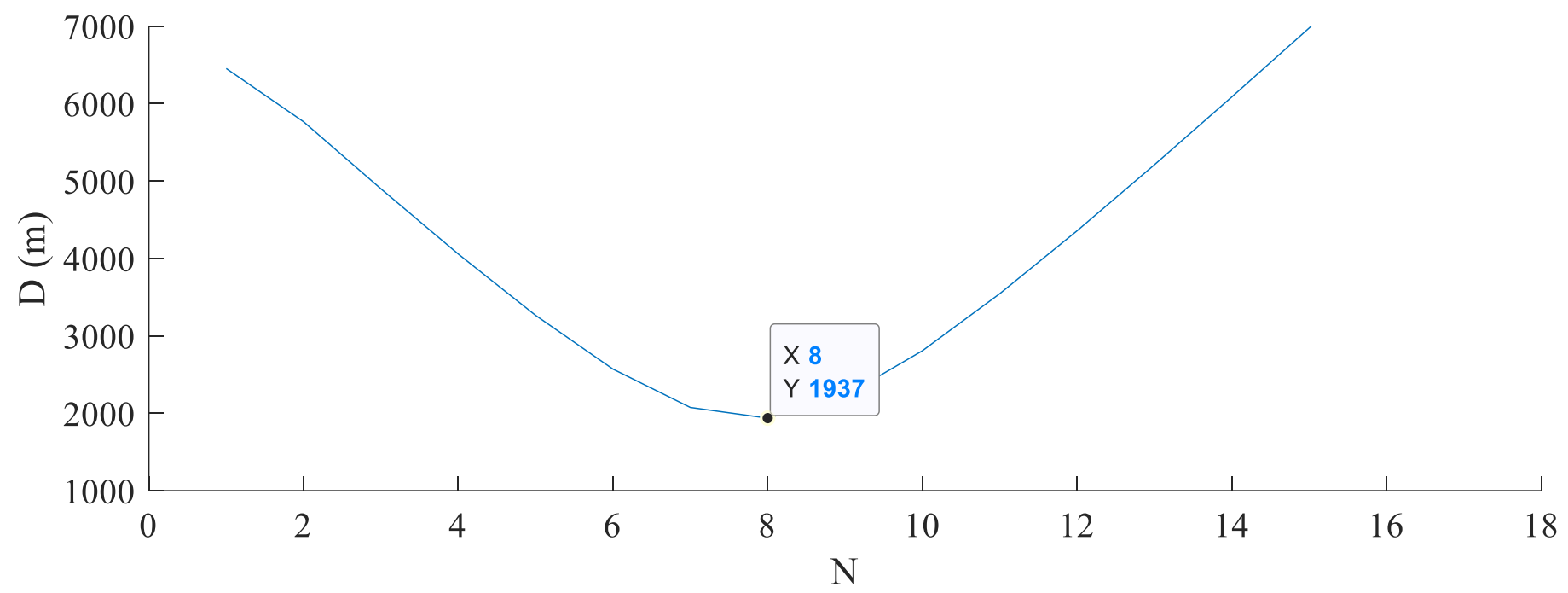

**Figure 29.** Distance between the USV and the obstacle in situation 2.

The third situation is the most complex and the process has two stages. It can be seen in Figure 30, the first stage in Figure 30 (a) - (c). Because there would be a collision at the intersection point, the USV has to steer in the beginning, and Figure 30 (b) is the critical state. After the obstacle being passed (Figure 30 (c)), the goal for the USV is to get to the destination, so the second stage in Figure 30 (c) - (e). From the perspective of the whole process, the USV makes the steering judgment in advance to avoid the dynamic obstacle. In fact, according to the proposed dynamic obstacle avoidance strategy in Section 4.5, there is a virtual obstacle at the intersection point (Figure 30 (f), the red dotted circle). The USV bypasses this virtual obstacle. The changes of distance between the USV and the obstacle are shown in Figure 31. The minimum distance is 1777 m which is still larger than the sum of the two ship domain radius.

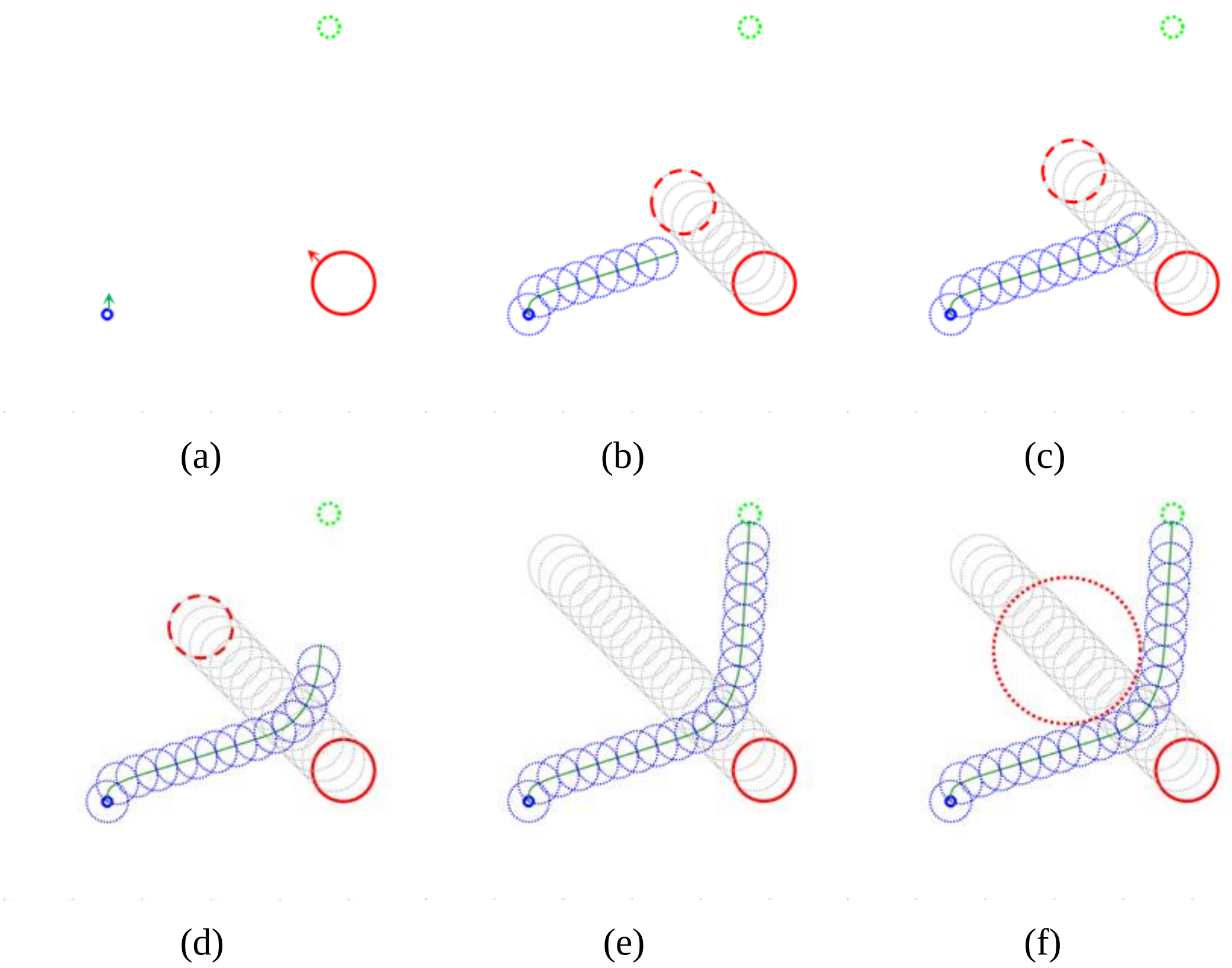

**Figure 30.** The process of dynamic obstacle avoidance (the most complex situation): (a) is the initial state; (b) is the critical state for the collision; (c) is the danger releasing state; (d) is the destination-driven state; (e) is the final state; (f) is the actual process that the USV bypass the virtual obstacle.

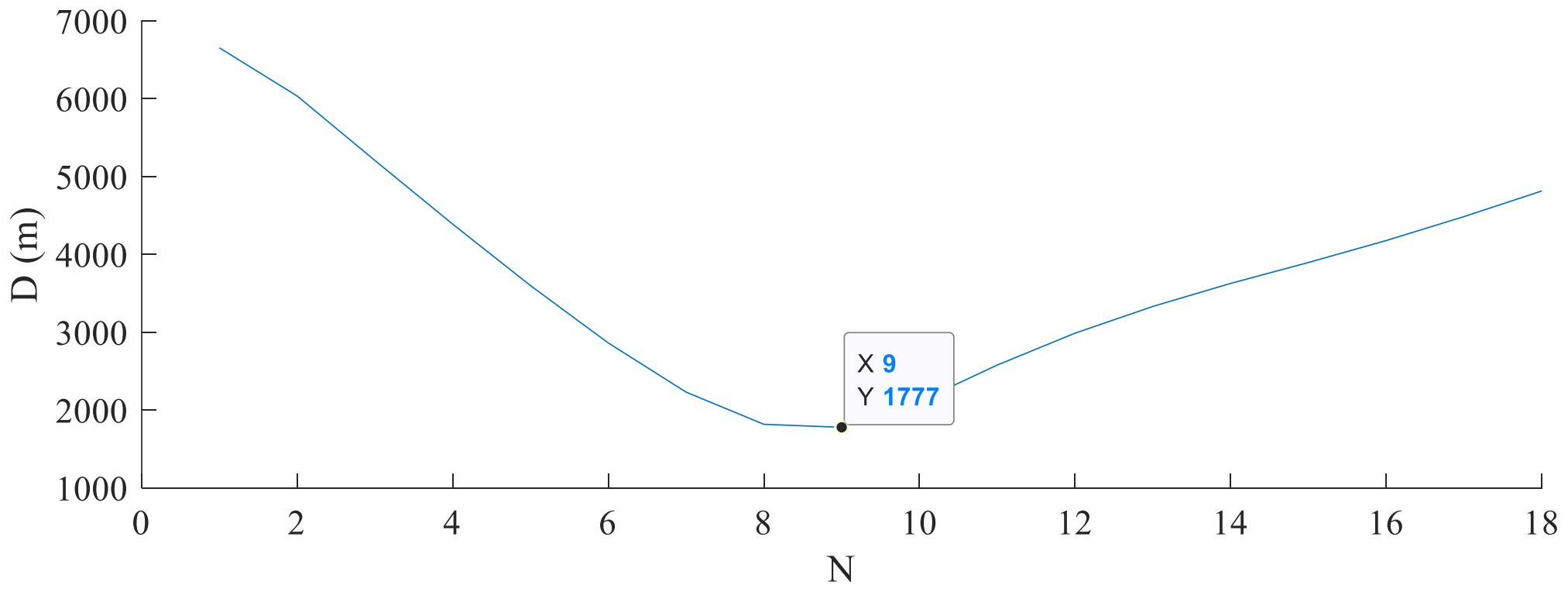

**Figure 31.** Distance between the USV and the obstacle in situation 3.

## 6. Conclusions

This paper proposes a motion planning method for USV obstacle avoidance based on CGTC, in which the Circle Grid is a way of map modeling. Compared to the traditional grid, the circle grid allows the USV candidate headings and waypoints to have a continuous change on the arc. And this characteristic is good at associating with a continuous motion curve. To solve the problem of space fully covering, this paper takes advantage of the structure of the circle grid tree which covers all the possible situations and has an easy way to calculate the positions. Besides, Trajectory Cell is a way of combining the motion curve with space search, which adds trajectory constraints into the process of obstacle avoidance, so this model contains trajectory producing and standardization rules. Furthermore, the trajectories are produced by the mathematic model of the USV, which can completely express the dynamic constraints. The standardization rules divide the trajectories into several curve segments and make sure these segments maintain continuous when they are spliced.

Based on the above model and rules, the obstacle avoidance method takes CGTC as the basic unit for USV motion planning and collision avoidance. The key to the collision avoidance method is to construct the relational function. This function associates the position and heading with the command rudder angle of the USV. And the heading can be determined by the geometrical relationship of the current position, final position, and obstacle states. So a proper USV trajectory can be produced in each process of heading selection. The experiments are carried out to evaluate the effectiveness of our proposed method, and the results indicate that the proposed algorithm can well avoid both static and dynamic obstacles compared with the relative methods considering USV safety, constraints of USV kinematics and dynamics.

In the future, the cooperative navigation and control for multi USVs might be considered to carry out complex tasks under uncertain environments.

## Acknowledgment

This work is supported by the National Natural Science Foundation of China (NSFC) through Grant No. 52001237.

## Reference

Almeida, C., Franco, T., Ferreira, H., Martins, A., Santos, R., Almeida, J. M., & Silva, E. (2009, May). Radar based collision detection developments on USV ROAZ II. In Oceans 2009-Europe (pp. 1-6), IEEE.

Belta, C., Isler, V., & Pappas, G. J. (2005). Discrete abstractions for robot motion planning and control in polygonal environments. IEEE Transactions on Robotics, 21(5), 864-874.

Bullo, F., & Lynch, K. M. (2001). Kinematic controllability for decoupled trajectory planning in underactuated mechanical systems. IEEE Transactions on Robotics and Automation, 17(4), 402-412.

Campbell, S., & Naeem, W. (2012). A rule-based heuristic method for COLREGS-compliant collision avoidance for an unmanned surface vehicle. IFAC proceedings volumes, 45(27), 386-391.

Choi, Y., Kim, D., Hwang, S., Kim, H., Kim, N., & Han, C. (2017). Dual-arm robot motion planning for collision avoidance using B-spline curve. International journal of precision engineering and manufacturing, 18(6), 835-843.

Chen, P., Huang, Y., Papadimitriou, E., Mou, J., & van Gelder, P. H. A. J. M. (2020a). An improved time discretized non-linear velocity obstacle method for multi-ship encounter detection. Ocean Engineering, 196, 106718.

Chen, P., Huang, Y., Papadimitriou, E., Mou, J., & van Gelder, P. (2020b). Global path planning for autonomous ship: A hybrid approach of Fast Marching Square and velocity obstacles methods. Ocean Engineering, 214, 107793.

Chen Y. (2016). A preliminary study on the optimized collaborative strategy and implementation of the unmanned surface vehicle. Jiangsu University of Science and Technology.

Dahl A. R. (2013). Path planning and guidance for marine surface vessels. Norwegian University of Science and Technology.

Du K., Mao Y., Xiang Z. (2015). A dynamic obstacle avoidance method for USV based on COLREGS. Ship Ocean Engineering, 44(3), 119-124.

Du, Z., Wen, Y., Xiao, C., Zhang, F., Huang, L., & Zhou, C. (2018). Motion planning for unmanned surface vehicle based on trajectory unit. Ocean Engineering, 151, 46-56.

Filliat, D., & Meyer, J. A. (2003). Map-based navigation in mobile robots: I. a review of localization strategies. Cognitive Systems Research, 4(4), 243-282.

Tan, C. Y., Huang, S., Tan, K. K., & Teo, R. S. H. (2020). Three dimensional collision avoidance for multi unmanned aerial vehicles using velocity obstacle. Journal of Intelligent & Robotic Systems, 97(1), 227-248.

Fujii, Y., & Tanaka, K. (1971). Traffic capacity. The Journal of navigation, 24(4), 543-552.

Gu, S., Zhou, C., Wen, Y., Zhong, X., Zhu, M., Xiao, C., & Du, Z. (2020). A motion planning method for unmanned surface vehicle in restricted waters. Proceedings of the Institution of Mechanical Engineers, Part M: Journal of Engineering for the Maritime Environment, 234(2), 332-345.

Gu, S., Zhou, C., Wen, Y., Xiao, C., Du, Z., & Huang, L. (2019). Path Search of Unmanned Surface Vehicle Based on Topological Location. Navigation of China, 42(02), 52-58.

Hansen, M. G., Jensen, T. K., Lehn-Schiøler, T., Melchild, K., Rasmussen, F. M., & Ennemark, F. (2013). Empirical ship domain based on AIS data. The Journal of Navigation, 66(6), 931-940.

Han, J., Cho, Y., Kim, J., Kim, J., Son, N. S., & Kim, S. Y. (2020). Autonomous collision detection and avoidance for ARAGON USV: Development and field tests. Journal of Field Robotics.

Huang, Y., Chen, L., Negenborn, R. R., & van Gelder, P. H. A. J. M. (2020a). A ship collision avoidance system for human-machine cooperation during collision avoidance. Ocean Engineering, 217, 107913.

Huang, Y., & van Gelder, P. H. A. J. M. (2020b). Collision risk measure for triggering evasive actions of maritime autonomous surface ships. Safety science, 127, 104708.

Jha, B., Chen, Z., & Shima, T. (2020). On shortest Dubins path via a circular boundary. Automatica, 121, 109192.

Kijima, K., Katsuno, T., Nakiri, Y., & Furukawa, Y. (1990). On the manoeuvring performance of a ship with theparameter of loading condition. Journal of the society of naval architects of Japan, 1990(168), 141-148.

Kim, H., Kim, D., Shin, J. U., Kim, H., & Myung, H. (2014). Angular rate-constrained path planning algorithm for unmanned surface vehicles. Ocean Engineering, 84, 37-44.

Lee, T. K., Baek, S. H., Choi, Y. H., & Oh, S. Y. (2011). Smooth coverage path planning and control of mobile robots based on high-resolution grid map representation. Robotics and Autonomous Systems, 59(10), 801-812.

Li, M., Mou, J., Chen, L., Huang, Y., & Chen, P. (2021). Comparison between the collision avoidance decision-making in theoretical research and navigation practices. Ocean Engineering, 108881.

Lyu, H., & Yin, Y. (2019). COLREGS-constrained real-time path planning for autonomous ships using modified artificial potential fields. The Journal of Navigation, 72(3), 588-608.

Shanmugavel, M., Tsourdos, A., White, B., & Żbikowski, R. (2010). Co-operative path planning of multiple UAVs using Dubins paths with clothoid arcs. Control Engineering Practice, 18(9), 1084-1092.

Mina, T., Singh, Y., & Min, B. C. (2019, October). A Novel Double Layered Weighted Potential Field Framework for Multi-USV Navigation towards Dynamic Obstacle Avoidance in a Constrained Maritime Environment. In OCEANS 2019 MTS/IEEE SEATTLE (pp. 1-9). IEEE.

Mirjalili, S. (2019). Genetic algorithm. In Evolutionary algorithms and neural networks (pp. 43-55). Springer, Cham.

Niu, H., Ji, Z., Savvaris, A., & Tsourdos, A. (2020). Energy efficient path planning for Unmanned Surface Vehicle in spatially-temporally variant environment. Ocean Engineering, 196, 106766.

Ogawa, Oyama. (1997). MMG Report-I-Mathematical model of control movement. The Japan Shipbuilding Society.

Orozco-Rosas, U., Montiel, O., & Sepúlveda, R. (2019). Mobile robot path planning using membrane evolutionary artificial potential field. Applied Soft Computing, 77, 236-251.

Singh, Y., 2019. Cooperative swarm optimization of unmanned surface vehicles (Doctoral dissertation, University of Plymouth).

Song, R., Liu, Y., & Bucknall, R. (2019). Smoothed A* algorithm for practical unmanned surface vehicle path planning. Applied Ocean Research, 83, 9-20.

Sun X. (2016). Research on the real-time path planning system of unmanned surface vehicle. Dalian Maritime University.

Svec, P., Thakur, A., Shah, B. C., & Gupta, S. K. (2012, August). USV trajectory planning for time varying motion goals in an environment with obstacles. In International Design Engineering Technical Conferences and Computers and Information in Engineering Conference (Vol. 45035, pp. 1297-1306). American Society of Mechanical Engineers.

Thrun, S., Gutmann, J. S., Fox, D., Burgard, W., & Kuipers, B. (1998, July). Integrating topological and metric maps for mobile robot navigation: A statistical approach. In AAAI/IAAI (pp. 989-995).

Wang, J., & Meng, M. Q. H. (2020). Optimal path planning using generalized Voronoi graph and multiple potential functions. IEEE Transactions on Industrial Electronics.

Wei-Dong, Z., Xiao-Cheng, L., & Peng, H. (2020). Progress and challenges of overwater unmanned systems. Acta Automatica Sinica, 46(5), 847-857.

Woo, J., & Kim, N. (2020). Collision avoidance for an unmanned surface vehicle using deep reinforcement learning. Ocean Engineering, 199, 107001.

Wu, K., Esfahani, M. A., Yuan, S., & Wang, H. (2019). Tdpp-net: Achieving three-dimensional path planning via a deep neural network architecture. Neurocomputing, 357, 151-162.

Yahja, A., Stentz, A., Singh, S., & Brumitt, B. L. (1998, May). Framed-quadtree path planning for mobile robots operating in sparse environments. In Proceedings. 1998 IEEE International Conference on Robotics and Automation (Cat. No. 98CH36146) (Vol. 1, pp. 650-655). IEEE.

Yang, J. M., Tseng, C. M., & Tseng, P. S. (2015). Path planning on satellite images for unmanned surface vehicles. International Journal of Naval Architecture and Ocean Engineering, 7(1), 87-99.

Kuwata, Y., Wolf, M. T., Zarzhitsky, D., & Huntsberger, T. L. (2011, September). Safe maritime navigation with COLREGS using velocity obstacles. In 2011 IEEE/RSJ International Conference on Intelligent Robots and Systems (pp. 4728-4734). IEEE.

Zhuang, J. Y., Wan, L., Liao, Y. L., & Sun, H. B. (2011). Global path planning of unmanned surface vehicle based on electronic chart. Computer Science, 38(9), 211.

Zhou, C., Gu, S., Wen, Y., Du, Z., Xiao, C., Huang, L., & Zhu, M. (2020a). The review unmanned surface vehicle path planning: Based on multi-modality constraint. Ocean Engineering, 200, 107043.

Zhou, C., Gu, S., Wen, Y., Du, Z., Xiao, C., Huang, L., & Zhu, M. (2020b). Motion planning for an unmanned surface vehicle based on topological position maps. Ocean Engineering, 198, 106798.